\documentclass[runningheads]{llncs}

\usepackage[mobile]{eccv}

\usepackage{eccvabbrv}

\usepackage{graphicx}

\usepackage[accsupp]{axessibility}  %

\usepackage{hyperref}

\usepackage{lipsum}
\usepackage{moresize}
\usepackage{multirow}
\usepackage[capitalize]{cleveref}
\usepackage{wrapfig}
\usepackage{amsmath}
\usepackage{amssymb}
\usepackage{booktabs}
\usepackage{cases}
\usepackage{algorithm}
\usepackage{algpseudocode}
\usepackage{xcolor}

\usepackage{mathtools}

\usepackage{overpic}
\usepackage{rotating}
\usepackage{cancel}
\usepackage{subcaption}
\usepackage{pgfplotstable} %
\usepackage{tikz}
\usepackage{tikz-cd}
\usepackage{colortbl}
\usepackage{cancel}
\usepackage{multibib}
\usepackage{afterpage} %

\makeatletter
\AddToHook{cmd/appendix/before}{\def\cref@section@alias{appendix}}
\makeatother

\crefname{section}{Sec.}{Secs.}
\Crefname{section}{Section}{Sections}
\Crefname{table}{Table}{Tables}
\crefname{table}{Tab.}{Tabs.}
\Crefname{append}{Appendix}{Appendixs}
\crefname{append}{Append.}{Appends.}
\Crefname{subfigure}{Figure}{Figures}
\crefname{subfigure}{Fig.}{Figs.}

\usepackage{orcidlink}

\newcommand{\method}[1]{{CRD}}

\begin{document}

\title{Diffusion Reinforcement Learning via \\Centered Reward Distillation}

\author{Yuanzhi Zhu$^{1}$
\quad
Xi Wang$^1$
\quad
Stéphane Lathuilière$^{2}$
\quad
Vicky Kalogeiton$^{1}$
}

\authorrunning{Y.~Zhu et al.}

\institute{LIX, École Polytechnique, CNRS, IPP \and
Inria at Univ. Grenoble Alpes, CNRS, LJK
}

\maketitle

\begin{abstract}

Diffusion and flow models achieve State-Of-The-Art (SOTA) generative performance, yet many practically important behaviors such as fine-grained prompt fidelity, compositional correctness, and text rendering are weakly specified by score or flow matching pretraining objectives. 
Reinforcement Learning (RL) fine-tuning with external, black-box rewards is a natural remedy, but diffusion RL is often brittle. 
Trajectory-based methods incur high memory cost and high-variance gradient estimates; forward-process approaches converge faster but can suffer from distribution drift, and hence reward hacking.
In this work, we present \textbf{Centered Reward Distillation (\method{})}, a diffusion RL framework derived from KL-regularized reward maximization built on forward-process-based fine-tuning.
The key insight is that the intractable normalizing constant cancels under \emph{within-prompt centering}, yielding a well-posed reward-matching objective.
To enable reliable text-to-image fine-tuning, we introduce techniques that explicitly control distribution drift: (\textit{i}) decoupling the sampler from the moving reference to prevent ratio-signal collapse, (\textit{ii}) KL anchoring to a CFG-guided pretrained model to control long-run drift and align with the inference-time semantics of the pre-trained model, and (\textit{iii}) reward-adaptive KL strength to accelerate early learning under large KL regularization while reducing late-stage exploitation of reward-model loopholes.  
Experiments on text-to-image post-training with \texttt{GenEval} and \texttt{OCR} rewards show that \method{} achieves competitive SOTA reward optimization results with fast convergence and reduced reward hacking, as validated on unseen preference metrics.

\end{abstract}

\begin{figure}[t!]
    \centering
    \includegraphics[width=0.95\linewidth]{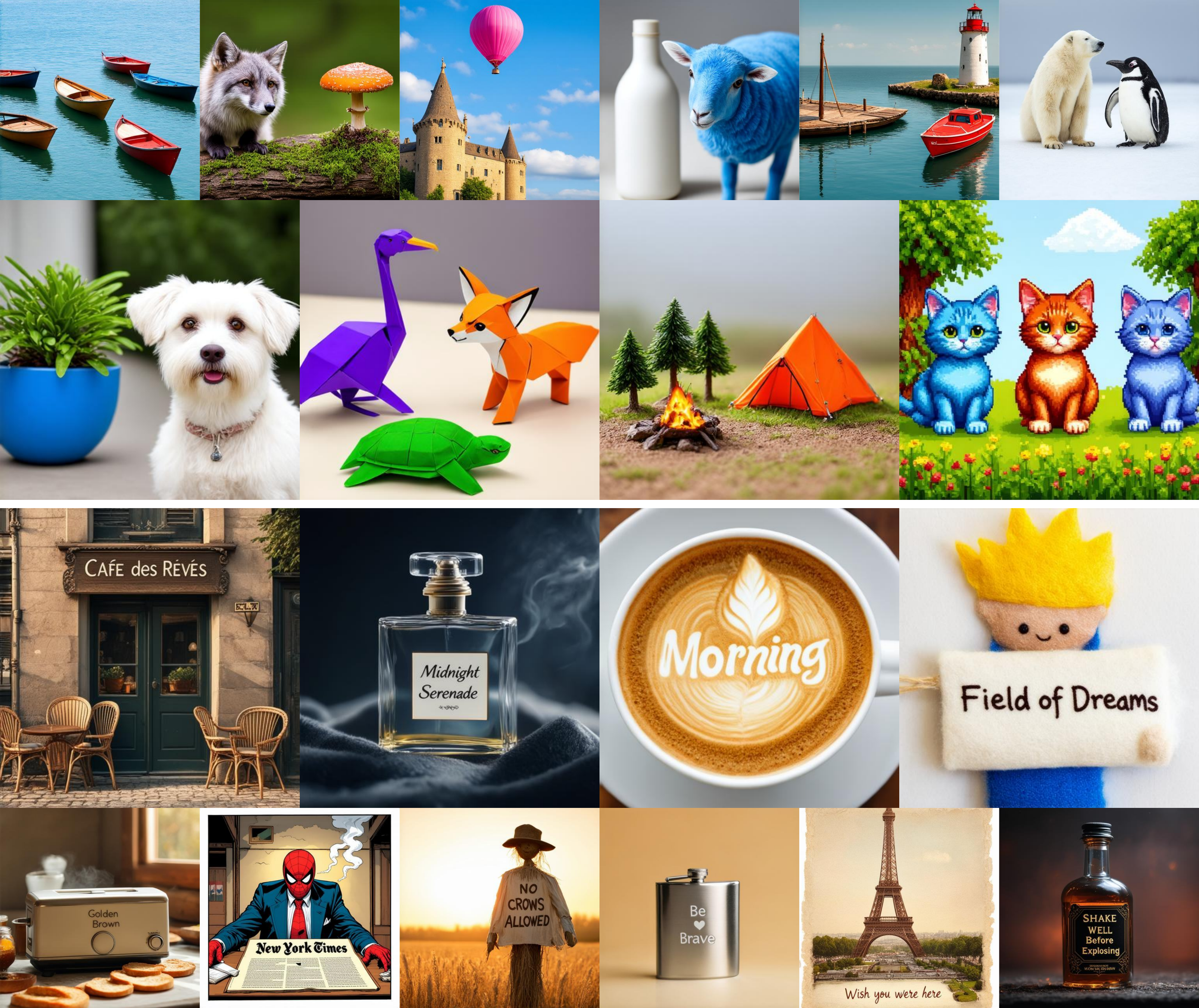}
    \caption{Qualitative results produced by our RL fine-tuned SD3.5M \cite{esser2024scaling} model with \texttt{GenEval}~\cite{ghosh2023geneval} reward (top) and \texttt{OCR}~\cite{chen2023textdiffuser} reward (bottom).
    }
    \vspace{-15pt}
    \label{fig:teaser}
\end{figure}

\section{Introduction}

Diffusion and flow models \cite{sohl2015deep,ho2020denoising,song2019generative,song2020score,vahdat2021score,elucidating,lipman2022flow,albergo2023stochastic,liu2022flow,liu2025blessing} have become a cornerstone of modern generative modeling, achieving state-of-the-art performance across modalities and tasks \cite{ho2022video,rombach2022high,peebles2023scalable,dufour2024don,wang2025akira,degeorge2025farimagenettexttoimagegeneration,boudier2025dipsy,courant2025pulp,esser2024scaling,wan2025wan}.
Yet many behaviors that matter most in practice, such as aesthetic quality, fine-grained prompt fidelity, and legible text rendering, are only weakly specified by denoising score matching \cite{vincent2011connection,song2019generative} objectives on static datasets, and are therefore not reliably induced during pretraining.
Closing this gap calls for \emph{post-training} strategies that optimize models with respect to external signals. 
Reinforcement Learning (RL) is particularly well-suited to this end: rather than matching a fixed data distribution, the model is explicitly optimized toward downstream objectives using rewards derived from human preference models \cite{wu2023human,ma2025hpsv3,xu2023imagereward,schuhmann2022aesthetics}, 
vision-language evaluators \cite{hessel2021clipscore}, or task-specific metrics 
\cite{chen2023textdiffuser,ghosh2023geneval}.  %

Diffusion RL aims to \emph{post-train} a pretrained model to maximize external reward signals while preventing excessive drift from the original distribution. Early approaches either operated off-policy with expensive data collection process, or required differentiable rewards (\emph{e.g.}, \texttt{CLIPScore} \cite{hessel2021clipscore}) to backpropagate through the denoising process~\cite{peng2019advantage,clark2023directly,fan2023optimizing,black2023training,lee2023aligning,yuan2024self}, limiting their applicability to reward signals that are non-differentiable (\emph{e.g.}, \texttt{OCR} accuracy, object detection metrics) prohibitively expensive 
to differentiate through (\emph{e.g.}, VLM-based evaluators).
To address these, recent diffusion RL work has pursued two main directions.

First, GRPO-style methods \cite{shao2024deepseekmath,guo2025deepseek} (e.g., Flow-GRPO \cite{liu2025flow,xue2025dancegrpo} and follow-ups \cite{li2025branchgrpo,ding2025treegrpo,deng2026densegrpo,liu2025diversegrpo,chen2025superflow}) view denoising as an explicit Markov Decision Process (MDP): each denoising step is treated as an action that transforms the current noisy latent into a less noisy one, and the full SDE sampling trajectory plays the role of an RL rollout.
This perspective is attractive as it yields an explicit sequential decision problem with well-defined intermediate states and actions, enabling principled credit assignment across denoising steps and importing variance-reduction and off-policy tools (advantages/baselines, importance ratios, trust-region style updates). 
These same structural benefits also align closely with the RLHF techniques developed for autoregressive LLMs \cite{guo2025deepseek,yu2025dapo,liu2025understanding,yue2025does}, making GRPO-style formulations a natural extension of that toolkit.
However, this first direction, due to its rollout-based MDP formulation, typically requires storing the full SDE trajectory and suffers from high-variance training signals, imposing high memory and compute overhead that slows convergence, which is particularly prohibitive when post-training large-scale diffusion models \cite{wan2025wan}.

A second line of work \cite{xue2025advantage,zheng2025diffusionnft,luo2025reinforcing,wang2026gdro,choi2026rethinking,chen2025towards,ou2026diffusion} argues that diffusion RL can be formulated more directly via the \emph{forward process}. The key idea is to \emph{decouple sampling from training states} \cite{zheng2025diffusionnft}: 
clean samples are first generated from the current model, noisy latents are then obtained via the forward diffusion process, and training proceeds on these synthetically noised states.
This yields objectives that resemble \emph{advantage-weighted maximum likelihood}, intuitively, high-reward samples are upweighted and low-reward samples are downweighted.
Compared to the Flow-GRPO direction, forward-process objectives are typically simpler to implement and substantially more efficient, with lower-variance gradients that often converge faster \cite{xue2025advantage}.
However, this direction introduces its own challenge: a fixed reference model becomes a poor surrogate as training progresses, making the distance between current and reference model increasingly large and training samples off-distribution.
To mitigate this, these methods maintain a \emph{moving reference} that is periodically updated toward the current model; but if it tracks too aggressively, it will enable iterative drift \cite{xue2025advantage,zheng2025diffusionnft}.
In practice, such drift can amplify reward hacking by encouraging exploitation of reward imperfections.

In this work, we revisit the forward-process paradigm through the lens of KL-regularized reward maximization, where the optimal solution is a reward-tilted \cite{peters2007reinforcement,peng2019advantage,korbak2022reinforcement,go2023aligning,rafailov2023direct,uehara2025inference,kim2025test,pachebat2025iterative,sabour2025test,potaptchik2025tilt} version of a reference model. This perspective implies that external rewards correspond to a scaled log-density ratio between the optimal fine-tuned model and the reference model, but only \emph{up to an unknown, prompt-dependent normalizer}. Because this normalizer is intractable, directly regressing model likelihood ratios toward absolute rewards is ill-posed \cite{mao2024don,gao2024rebel,fisch2024robust,zhu2025flowrl,matrenok2025quantile}.
Our key insight is that the normalizer cancels exactly under \emph{within-prompt centering}: for a group of samples drawn under the same prompt, the unknown normalizer term is identical across samples and disappears when rewards are centered within the group.
Building on this observation, we propose \textbf{Centered Reward Distillation (\method{})}, a set of well-posed reward-matching objectives for diffusion RL. \method{} trains the diffusion model so that its implicit log-density ratio (approximated via a diffusion Evidence Lower BOund surrogate, ELBO) matches \emph{centered} external rewards within each prompt group. This yields a unifying view that recovers prior works in LLM such as reward-distillation methods \cite{mao2024don,gao2024rebel,fisch2024robust} and GVPO objective \cite{zhang2025gvpo} as special cases, and naturally extends to a ratio-based variant connected to InfoNCA \cite{chen2024noise}.

To make \method{} reliable for {moving reference RL fine-tuning},
we introduce practical stabilization techniques that explicitly mitigate reward hacking by controlling distribution drift. 
First, we decouple the model used to generate samples from the \emph{moving reference} used in the log-ratio objective, which prevents training instabilities when the reference drifts too close to the current model. 
Second, to control long-run drift, we add a KL penalty that anchors the current model to a \emph{fixed} initial reference (the pretrained model). 
Importantly, because many forward-process setups train and sample the current model \emph{without} Classifier-Free Guidance (CFG) \cite{ho2022classifier,wang2024analysis,kynkaanniemi2024applying} for efficiency, we adopt the \emph{CFG-guided} version of the initial reference model, aligning regularization with inference-time semantics (and reducing to CFG distillation when the RL signal is absent). 
Finally, we use a simple reward-adaptive scaling of the anchoring strength to accelerate early learning while preventing late-stage exploitation of reward-model loopholes.

We evaluate \method{} on text-to-image RL fine-tuning with \texttt{GenEval} and \texttt{OCR}-based rewards, comparing against representative Flow-GRPO-style and forward-process baselines. 
Overall, our approach achieves competitive reward optimization with faster and more stable training, and reduced reward hacking behavior.

Our contributions can be summarized as follows: 
\begin{itemize}
    \item We introduce \textbf{Centered Reward Distillation (\method{})}, a within-prompt centered reward-matching framework for diffusion RL that removes the unknown prompt-dependent normalizer, yielding a well-posed objective; the framework recovers prior reward distillation methods and GVPO-style objectives as special cases and admits a ratio-based variant connected to InfoNCA under appropriate parameterization.
    \item We propose practical techniques for stable RL fine-tuning: decoupled sampling versus moving reference, KL anchoring to a \emph{CFG}-guided fixed reference, and reward-adaptive KL strength to reduce drift and reward hacking 
    \item We provide text-to-image experiments on \texttt{GenEval} and \texttt{OCR} rewards demonstrating competitive performance against recent diffusion RL baselines, with improved stability and reduced reward hacking.
\end{itemize}

\section{Related Works}
{Progress in diffusion and flow-based generative modeling has motivated works that apply RL as a \emph{post-training} stage to better align generations with human preferences or task-specific reward signals \cite{peng2019advantage,clark2023directly,fan2023optimizing,black2023training,lee2023aligning,yuan2024self}.}

\subsection{GRPO in Visual Generation}
{Recently, methods leveraging \emph{Group Relative Policy Optimization} (GRPO) \cite{shao2024deepseekmath,guo2025deepseek} have shown strong alignment performance in diffusion and flow-based models \cite{liu2025flow, xue2025dancegrpo}.}
Meanwhile, a recurring challenge in diffusion/flow GRPO is the denoising process itself: learning signals and estimator variance are highly timestep-dependent, while practical training also hinges on rollout fidelity, sampling efficiency, and the choice of reward objective. Existing methods address these issues through several complementary mechanisms:
\vspace{-5pt}
\begin{enumerate}
    \item \textbf{Trajectory structuring for exploration and credit assignment.}
    These methods modify rollout topology (branching/trees/chunks) to broaden exploration and improve credit assignment along denoising:
    TempFlow-GRPO \cite{he2025tempflow}; Branch-GRPO \cite{li2025branchgrpo}, TreeGRPO \cite{ding2025treegrpo}; Chunk-GRPO \cite{luo2025sample}.
    The trade-off is added algorithmic complexity and sensitivity to design choices (e.g., branching depth, chunk size).

    \item \textbf{Dense rewards and ratio/regularization stabilizers.}
    This line reshapes timestep-wise learning signals and stabilizes per-step weighting via dense rewards or ratio/regularization control:
    Dense-GRPO \cite{deng2026densegrpo}, GARDO \cite{he2025gardo}, GRPO-Guard (RatioNorm) \cite{wang2025grpo}.
    Gains depend on the quality of dense reward proxies and careful calibration of normalization/regularization.

    \item \textbf{Sampling efficiency and rollout fidelity.}
    These works improve the compute--quality trade-off by increasing group-sampling efficiency and reducing solver-induced artifacts:
    SuperFlow \cite{chen2025superflow}, Flow-CPS \cite{wang2025coefficients}, E-GRPO \cite{zhang2026grpo}, Neighbor GRPO \cite{he2025neighbor}.
    They are largely orthogonal to reward design but can remain brittle under sparse or misaligned rewards.

    \item \textbf{Richer reward definitions and objectives.}
    Recent variants broaden supervision beyond a single prompt-level scalar (e.g., reference-image rewards, diversity objectives, prompt refinement, or multi-granularity advantages):
    Adv-GRPO \cite{mao2025image}, DiverseGRPO \cite{liu2025diversegrpo}, PromptRL \cite{wang2026promptrl}, Granular-GRPO \cite{zhou2025fine}.
    While often more informative, they introduce extra assumptions/modules and can increase sensitivity to reward/model misspecification.
\end{enumerate}
\vspace{-3pt}

Despite these advances, a fundamental limitation remains: most of these methods optimize through the \emph{backward-process} MDP formulation, introducing inherent challenges in timestep-dependent variance, rollout overhead, and credit assignment that ultimately limit training efficiency.

\subsection{Diffusion RL Based on the Forward Process}
In contrast to Flow-GRPO and subsequent methods that explicitly model the diffusion sampling trajectory as an MDP and derive policy-gradient style updates under that assumption, a recent line of work argues that diffusion RL can be formulated more directly using the \emph{forward-process} \cite{xue2025advantage,zheng2025diffusionnft,luo2025reinforcing,wang2026gdro,choi2026rethinking,chen2025towards,zhu2025enhancing}. 
{
The key idea is to decouple data sampling and model training: in the sampling phase, the model generates clean samples without storing intermediate noisy states; in the training phase, the loss is computed on noisy versions of the samples using the forward diffusion process.}
{Experiments and analyses from these works suggest that forward-process-based methods achieve significantly faster training compared to backward-process-based methods like Flow-GRPO.}

GPO \cite{chen2025towards} is the first attempt at this perspective. It employs an ELBO-based likelihood estimator for diffusion models and shows that the resulting objective can be interpreted as an \emph{advantage-weighted log-likelihood}, connecting diffusion RL to weighted maximum likelihood updates. 
{Concurrently, AWM \cite{xue2025advantage} analyzes the variance properties of Flow-GRPO-style estimators and proposes a forward-process algorithm, \emph{Advantage Weighted Matching}, to mitigate variance amplification, demonstrating that it significantly accelerates training.}
An important conceptual link is that the GPO objective can be viewed as a first-order approximation of AWM {objective}; the approximation error becomes negligible when the current model remains close to the reference (old) model \cite{xue2025advantage,chen2025towards}.

DiffusionNFT \cite{zheng2025diffusionnft} also adopts a forward-process RL formulation and advocates using faster ODE-based samplers during training to improve efficiency. Unlike approaches that explicitly rely on an ELBO likelihood estimator to construct loss function, DiffusionNFT uniquely derives an update that targets an optimal velocity direction that steers samples toward high-reward regions. 
Despite this difference in derivation, DiffusionNFT remains conceptually aligned with the forward-process family in that it targets the same optimal solution.

DGPO \cite{luo2025reinforcing} further develops the forward-process paradigm by learning from \emph{group-level} preferences, exploiting relative information among samples within the same group and empirically demonstrating efficient optimization and stable convergence. 
GDRO \cite{wang2026gdro} reframes the log-likelihood ratio and advantage in terms of probability and derives a cross-entropy training objective; the resulting loss is closely related to InfoNCA proposed in \cite{chen2024noise}. 
Finally, {Choi et al.} \cite{choi2026rethinking} provides a systematic study of design choices in diffusion RL and empirically validates the central contribution from the ELBO-based estimator in diffusion RL training.

{Our method falls within this forward-process RL family and inherits its computational efficiency, while introducing a new class of objectives and practical techniques that prevent reward hacking without sacrificing training efficiency.}

\vspace{-10pt}
\section{Diffusion RL using Reward Distillation}
\label{sec:diffusion_reward_distil}

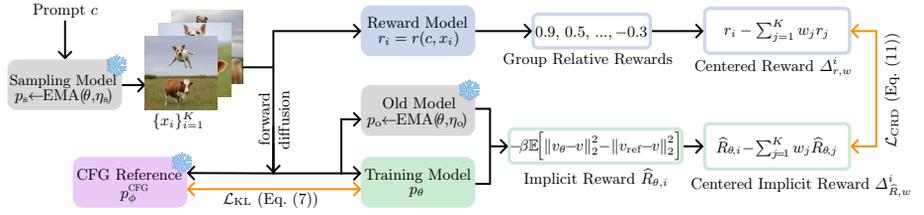
\begin{figure}[!t]
    \centering
    \begin{overpic}[width=0.98\linewidth]{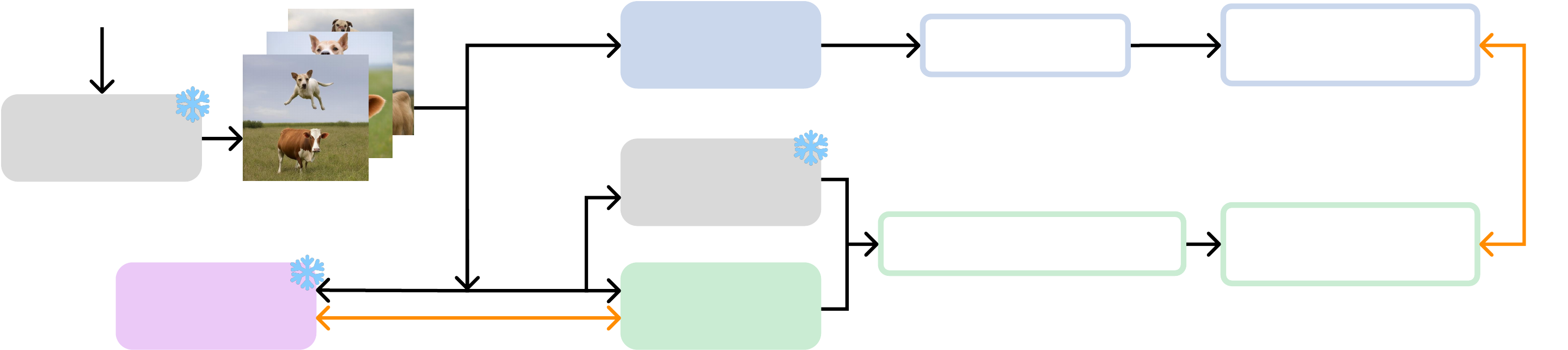}%
    \put(2.9,21.2){\color{black}\scalebox{0.65}{Prompt $c$}}
    \put(0.32,13.5){\color{black}\scalebox{0.62}{Sampling Model}}
    \put(0.9,11.8){\color{black}\scalebox{0.65}{$p_{\mathrm{s}} \!\!\leftarrow\!\! \mathrm{EMA}(\!\theta,\!\eta_{\mathrm{s}}\!)$}}
    \put(16.3,9.0){\color{black}\scalebox{0.65}{$\{x_i\}_{i=1}^K$}}
    \put(7.8,3.1){\color{black}\scalebox{0.65}{CFG Reference}}
    \put(12.7,1.1){\color{black}\scalebox{0.65}{$p_\phi^{\mathrm{\scalebox{0.6}{CFG}}}$}}
    \put(28.2,6.8){\rotatebox{90}{\color{black}\scalebox{0.65}{forward}}}
    \put(30.4,6.39){\rotatebox{90}{\color{black}\scalebox{0.65}{diffusion}}}
    \put(24.2,0.45){\color{black}\scalebox{0.65}{$\mathcal{L}_\mathrm{KL}$ (\cref{eq:rwrd_loss})}}
    \put(40.3,19.8){\color{black}\scalebox{0.65}{Reward Model}}
    \put(41.5,18.1){\color{black}\scalebox{0.65}{$r_i=r(c,x_i)$}}
    \put(42.0,10.9){\color{black}\scalebox{0.65}{Old Model}}
    \put(40.3,9.1){\color{black}\scalebox{0.65}{$p_{\mathrm{o}} \!\!\leftarrow\!\! \mathrm{EMA}(\!\theta,\!\eta_{\mathrm{o}}\!)$}}
    \put(40.,3.1){\color{black}\scalebox{0.65}{Training Model}}
    \put(45.0,1.5){\color{black}\scalebox{0.65}{$p_{\theta}$}}
    \put(59.1,18.8){\color{black}\scalebox{0.65}{$0.9$, $0.5$, ..., \!${-}0.3$}}
    \put(55.5,15.8){\color{black}\scalebox{0.65}{Group Relative Rewards}}
    \put(56.5,6.3){\color{black}\scalebox{0.65}{$\!{-}\!\beta \mathbb{E}\! \Big[\! \left\|v_\theta\!\!-\!\!v\right\|_2^2 \!{-}\! \left\|v_{\mathrm{ref}}\!\!-\!\!v\right\|_2^2 \!\Big]$}}
    \put(57.8,2.6){\color{black}\scalebox{0.65}{Implicit Reward $\widehat{R}_{\theta,i}$}}
    \put(80.2,19.0){\color{black}\scalebox{0.65}{$r_i-\sum_{j=1}^K w_j r_j$}}
    \put(76.9,15.0){\color{black}\scalebox{0.65}{Centered Reward $\Delta_{r,w}^i$}}
    \put(79.3,6.1){\color{black}\scalebox{0.65}{$\widehat{R}_{\theta\!,i}\!-\!\sum_{j\!=\!1}^K w_j\widehat{R}_{\theta\!,j}$}}
    \put(76.5,2.1){\color{black}\scalebox{0.65}{Centered Implicit Reward $\Delta_{\widehat{R},w}^i$}}
    \put(98.2,6.0){\rotatebox{90}{\color{black}\scalebox{0.65}{ $\mathcal{L}_\mathrm{\method{}}$ (\cref{eq:KL_loss})}}}
    \end{overpic}
    \vspace{-2pt}
    \caption{
    For each prompt, $K$ samples $\{x_i\}_{i=1}^K$ are generated from the sampling model $p_{\mathrm{samp}}$ ($p_s$). A reward model produces external rewards $r(c,x_i)$, and implicit model rewards $\widehat{R}_{\theta,i}$ are estimated via diffusion ELBO differences between the current model $p_\theta$ and a moving reference $p_{\mathrm{old}}$ ($p_o$). Within-prompt centering yields $\Delta_{r,w}^i$ and $\Delta_{\widehat R,w}^i$, cancelling the prompt-dependent normalizer and enabling a well-posed matching objective. 
    A initial KL penalty with respect to the fixed CFG-guided pretrained model $p_\phi^{\mathrm{\scalebox{0.8}{CFG}}}$ is imposed to prevent reward hacking. 
    }
    \vspace{-15pt}
    \label{fig:pipeline}
\end{figure}

{
We first introduce the KL-regularized formulation of diffusion RL and show that rewards correspond to a log-density ratio only up to an unknown prompt-dependent normalizer (\cref{sec:method_back}).  
We then present \textbf{Centered Reward Distillation (\method{})}, which removes this normalizer via within-prompt centering and yields a family of reward-matching objectives (\cref{sec:rwrd}).  
Finally, we describe practical techniques that mitigate reward hacking under distribution drift (\cref{sec:cfg_ref_kl}).
The overall training pipeline is illustrated in \cref{fig:pipeline} and the algorithm is summarized in \cref{alg:rl}.
}

\subsection{Background}
\label{sec:method_back}
\paragraph{Setup and goal.}
{Given access to a fixed external reward $r(c,x)$ (e.g., from a reward model) and a pretrained diffusion model with parameters $\phi$,
the goal of diffusion RL is to fine-tune a copy of this model with parameters $\theta$ 
to maximize the expected reward $r(c,x)$ of its generated samples,
while keeping the distribution $p_\theta(x | c)$ close to a reference distribution $p_{\mathrm{ref}}(x | c)$ (which is usually the pretrained model $p_\phi(x|c)$) for each prompt condition $c$.}

\paragraph{KL-regularized form and the unknown normalizer.}
Under the standard KL-regularized reward maximization objective, the optimal distribution satisfies \cite{peters2007reinforcement,peng2019advantage,korbak2022reinforcement,go2023aligning,rafailov2023direct}

\begin{equation}
\label{eq:rwrd_boltzmann}
p_{\theta^*}(x\mid c)\;\propto\;p_{\mathrm{ref}}(x\mid c)\exp\!\left(\frac{r(c,x)}{\beta}\right), 
\end{equation}
which implies
\begin{equation}
\label{eq:rwrd_reward_density_ratio}
r(c,x)
\;=\;
\beta \log \frac{p_{\theta^*}(x\mid c)}{p_{\mathrm{ref}}(x\mid c)}
+\beta \log Z(c),
\end{equation}
where $\beta\!>\!0$ is the KL regularization strength and
$Z(c)\!=\!\int \! p_{\mathrm{ref}}(x| c)\! \exp(\frac{r(c,x)}{\beta})dx$ is a prompt-dependent normalizer.

\paragraph{Implicit model reward and diffusion-specific surrogate.}
The \emph{implicit model reward} is defined as the scaled log-density ratio:
\begin{equation}
\label{eq:rwrd_policy_reward}
R_\theta(c,x)\;\triangleq\;\beta \log \frac{p_\theta(x\mid c)}{p_{\mathrm{ref}}(x\mid c)}.
\end{equation}
For diffusion models, the exact log-density ratio in \cref{eq:rwrd_policy_reward} is intractable.
Following \cite{wallace2024diffusion,mcallister2025flow,ren2024diffusion,choi2026rethinking}, we use the diffusion ELBO surrogate to estimate this term:
\begin{equation}
\label{eq:rwrd_policy_reward_elbo}
\widehat{R}_\theta(c,x)
\!\triangleq\!
-\beta\,
\mathbb{E}_{t,\epsilon} \! \left[w(t)(
\left\|v_\theta(x_t,t| c){-}v_\mathrm{target}\right\|_2^2
-
\left\|v_{\mathrm{ref}}(x_t,t| c){-}v_\mathrm{target}\right\|_2^2)
\right],
\end{equation}
where $t$ is sampled from a predefined timestep distribution, $\epsilon\sim\mathcal{N}(0,I)$, $w(t)$ is a time-dependent weighting which we set as 1 for simplicity, $x_t$ is the noisy latent produced by the forward diffusion process from clean data $x$ at time $t$, $v_\mathrm{target}(\cdot)=\epsilon-x$ denotes the corresponding target velocity and $v_\theta$ is the velocity predicted by the model $\theta$.
{We use $\widehat{R}_\theta$ as an estimator of $R_\theta$ throughout what follows.}
{Following previous work \cite{xue2025advantage,zheng2025diffusionnft}, we first transform the raw reward $r_{\mathrm{raw}}(c,x)$ using group normalization to get $r(c,x)$}.

{Crucially, since the normalizer $\beta\log Z(c)$ is intractable, directly regressing $\widehat{R}_\theta(c,x)$ onto absolute rewards $r(c,x)$ is ill-posed without knowing the normalizer based on \cref{eq:rwrd_reward_density_ratio} \cite{zhu2025flowrl,matrenok2025quantile}.}

\subsection{Centered Reward Distillation (\method{})}
\label{sec:rwrd}

\paragraph{Key insight: within-prompt centering (subtracting the weighted group reward mean) removes the normalizer $\beta \log Z(c)$.}
For each prompt $c$, we consider a group of $K$ samples $\{x_i\}_{i=1}^K$ drawn from a proposal distribution and their corresponding external rewards $r(c,x_i)$.
Recall from \cref{eq:rwrd_reward_density_ratio} that, under KL-regularized reward maximization, the reward decomposes as $r(c,x)= R_{\theta^*}(c,x) + \beta \log Z(c)$, where the normalizer term $\beta\log Z(c)$ depends only on the prompt $c$.
Consequently, for a fixed prompt $c$ and any weights $\{w_j\}_{j=1}^K$ satisfying $\sum_{j=1}^K w_j = 1$, the normalizer cancels under within-prompt centering:
\begin{equation}
\label{eq:rwrd_centering_identity}
r(c,x_i) - \sum_{j=1}^K w_j r(c,x_j)
\;=\;
R_{\theta^*}(c,x_i) - \sum_{j=1}^K w_j R_{\theta^*}(c,x_j).
\end{equation}
{This \textbf{centered matching} problem is well-posed regardless of the unknown normalizer, and directly motivates our objective: train the model so that its implicit reward $\widehat{R}_\theta(c,x)$ is consistent with the external rewards.}

\paragraph{Reward-weighted centering weights.} 
Within each prompt group, we define reward-weighted softmax weights:
$w_i(c,\{x_j\}_{j=1}^K;\tau)
\;\triangleq\;
\frac{\exp\!\left(r(c,x_i)/\tau\right)}{\sum_{j=1}^K \exp\!\left(r(c,x_j)/\tau\right)}$
{where $\tau>0$ is temperature that controls the sharpness of the probability distribution over the estimated rewards.} For brevity, we write $w_i \equiv w_i(c,\{x_j\}_{j=1}^K;\tau)$ throughout.
{We note that, }as $\tau\!\to\!\infty$, $w_i \to 1/K$ (uniform centering), whereas as $\tau\!\to\!0$, the weights concentrate on the highest-reward sample (max-anchored centering).

\paragraph{\method{} objective.}
Let $\rho(c,\{x_i\}_{i=1}^K)$ be a distribution over prompts and the corresponding within-prompt generation sets {with group size K} 
(sampled from a proposal such as the current model \cite{rafailov2023direct,ethayarajh2024kto,azar2024general}, a mixture with $p_{\mathrm{ref}}$ \cite{gorbatovski2024learn}, or a replay buffer \cite{xue2025advantage}). 
Motivated by the centering identity in \cref{eq:rwrd_centering_identity}, we define within-prompt centered residuals {for both the reward model and the log-density ratio estimator:}
\begin{equation}
\label{eq:rwrd_centered}
\Delta_{r,w}^i \;\triangleq\; r(c,x_i)-\sum_{j=1}^K w_j\, r(c,x_j),
\qquad
\Delta_{\widehat{R},w}^i \;\triangleq\; \widehat{R}_\theta(c,x_i)-\sum_{j=1}^K w_j\,\widehat{R}_\theta(c,x_j).
\end{equation}
The \textbf{\method{} loss} then matches centered external rewards to centered implicit model reward by minimizing
\begin{equation}
\label{eq:rwrd_loss}
\mathcal{L}_{\mathrm{\method{}}}^{(\tau)}(p_\theta;\rho)
=
\mathbb{E}_{\rho(c,\{x_i\})}\;
\frac{1}{K}\sum_{i=1}^K
\left(\Delta_{r,w}^i - \Delta_{\widehat{R},w}^i\right)^2.
\end{equation}

\paragraph{Connections to existing methods.}
\method{} can serve as a unifying framework that reinterprets several existing methods as special cases and admits natural extensions.
We show in \cref{supp:special_case} that two-sample reward distillation \cite{mao2024don,gao2024rebel,fisch2024robust} methods and GVPO \cite{zhang2025gvpo} can both be recovered as special cases of \method{}.
In addition, in \cref{supp:ratio_based} we introduce a ratio-based distillation variant of \method{}, and shows that InfoNCA loss proposed in \cite{chen2024noise} can be recovered as a special case under appropriate parameterization.

\paragraph{Training efficiency and stability: moving reference and decoupled sampling.}
{As training progresses, a fixed pretrained reference can become misaligned with the current model, yielding noisy log-density ratios and increasingly off-distribution samples. 
Following DiffusionNFT~\cite{zheng2025diffusionnft}, we maintain a \emph{moving reference} $p_{\mathrm{old}}$, held fixed within each epoch and updated at epoch end via Exponential Moving Average (EMA), so the reference and reference-sampled data remain close to $p_\theta$.
However, if $p_{\mathrm{old}}$ is updated too aggressively, it can drift too close to the current model $p_\theta$, collapsing the log-ratio signal in \cref{eq:rwrd_policy_reward} toward zero and destabilizing training. 
We therefore {further} decouple sampling from the log-ratio reference (see \cref{fig:pipeline}): samples are drawn from a separate EMA model $p_{\mathrm{samp}}$, while $p_{\mathrm{old}}$ is used only in the log-density ratio and updated with a slower EMA. 
This keeps data collection near on-policy while preserving a meaningful log-ratio signal.}

{
\begin{algorithm}[t]
    \scalebox{0.85}{
    \begin{minipage}{\linewidth}
    \begin{center}
    \caption{
    \method{} Distillation}
    \label{alg:rl}
    \begin{algorithmic}[1]
    \Require Pre-trained model ${\phi}$, reward model $r_\psi$, ODE solver $\mathrm{ODE}[\cdot]$, prompt dataset $\mathcal{D}_c$, $\beta_{\mathrm{old}}$ and $\beta_{\mathrm{init}}$, group size $K$, EMA decay $\eta_{\mathrm{old}}$ and $\eta_{\mathrm{samp}}$, temperature $\tau$
    \State{$\Delta\theta \!\leftarrow\! \text{initLoRA}(),\,\, \Delta\theta_{\mathrm{old}} \!\leftarrow\! \text{initLoRA}(),\,\, \Delta\theta_{\mathrm{samp}} \!\leftarrow\! \text{initLoRA()}$ \textcolor[rgb]{0.4,0.4,0.4}{{$\,\,\,$// init adapters}}}
    \State{$\theta \triangleq \phi + \Delta\theta, \,\, \theta_{\mathrm{old}} \triangleq \phi + \Delta\theta_{\mathrm{old}}, \,\, \theta_{\mathrm{samp}} \triangleq \phi + \Delta\theta_{\mathrm{samp}}$\textcolor[rgb]{0.4,0.4,0.4}{{$\,\,\,$// effective weights}}}
    \Repeat
        \State{\textcolor[rgb]{0,0.5,0}{\texttt{\textit{\#\#\# Sample prompts and group of data}}}}
        \State{Sample $c\sim \mathcal{D}_c$, $\{x_1^i\}_{i=1}^K\sim \mathcal{N}(0,1)$}
        \State{Generate a group of samples $\{x_0^i\}_{i=1}^K\leftarrow \mathrm{ODE}[v_{\theta_{\mathrm{samp}}}](\{x_1^i\}_{i=1}^K)$}
        
        \State{\textcolor[rgb]{0,0.5,0}{\texttt{\textit{\#\#\# Calculate $r(c,x)$}}}}
        \State{$r_{\mathrm{raw}}^i\ =  r_\psi(c,x_0^i)$\textcolor[rgb]{0.40,0.40,0.40}{$\,\,\,\qquad $// scaled to $[0,1]$}}
        \State{$r(c,x_0^i) = \frac{r_{\mathrm{raw}}^i-\mathrm{mean}(\{r_{\mathrm{raw}}\}^{1:K})}{\mathrm{std}(\{r_{\mathrm{raw}}\}^{1:K})}$}
        \State{\textcolor[rgb]{0,0.5,0}{\texttt{\textit{\#\#\# Estimate $R_\theta(c,x)\;\triangleq\;\beta \log \frac{p_\theta(x\mid c)}{p_{\mathrm{old}}(x\mid c)}$}}}}
        \State{Sample $t\!\sim\!\mathcal{U}[0,1]$ and noise $\epsilon\!\sim\! \mathcal{N}(0,1)$ for each $i$}
        \State{Calculate $x_t=t \epsilon+ (1-t) x_0$ and $v_\mathrm{target}=\epsilon-x_0$ for each $i$}
        \State{$R_\theta(c,x_0^i) \! \leftarrow\! -\beta_{\mathrm{old}} (||v_\theta\!-\!v_\mathrm{target}||^2 - ||v_{\theta_{\mathrm{old}}}\!-\!v_\mathrm{target}||^2)$ for each $i$ \textcolor[rgb]{0.40,0.40,0.40}{$\,$// ELBO estimator}}
        
        \State{\textcolor[rgb]{0,0.5,0}{\texttt{\textit{\#\#\# Loss calculation and update $\theta$}}}}
        \State{Calculate $\Delta_{r,w}^i$ and $\Delta_{\widehat{R},w}^i$ using $r(c,x_0^i)$ and $\widehat{R}_\theta(c,x_0^i)$ \textcolor[rgb]{0.40,0.40,0.40}{$\,\,\,$// \cref{eq:rwrd_centered}}}
        \State{Calculate $\mathcal{L}_{\mathrm{\method{}}}^{(\tau)} \;=\;\frac{1}{K}\sum_{i=1}^K \left(\Delta_{r,w}^i-\Delta_{\widehat{R},w}^i\right)^2$}
        \State{$\hat\beta^i_{\mathrm{init}} \leftarrow r_{\mathrm{raw}}^i\beta_{\mathrm{init}}$ \textcolor[rgb]{0.40,0.40,0.40}{$\,\,\,\,\,\,\,\,\,\,\,\,\qquad\qquad\qquad\qquad\qquad\qquad\qquad$// Adaptive reference KL}}
        \State{Calculate $\mathcal{L}_{\mathrm{KL}} \;=\;\frac{1}{K}\sum_{i=1}^K \hat\beta^i_{\mathrm{init}}||v_\theta(x_t^i,t,c) - v_{\phi}^{\mathrm{CFG}}(x_t^i,t,c)||^2$}
        \State{Update $\theta$ using gradient of $\mathcal{L}_{\mathrm{\method{}\_KL}} \;=\; \mathcal{L}_{\mathrm{\method{}}} + \mathcal{L}_{\mathrm{KL}}$}
        
        \State{\textcolor[rgb]{0,0.5,0}{\texttt{\textit{\#\#\# Update $\theta_{\mathrm{old}}$ and $\theta_{\mathrm{samp}}$}}}}
        \State{$\theta_{\mathrm{old}} \leftarrow \eta_{\mathrm{old}}\theta_{\mathrm{old}} + (1-\eta_{\mathrm{old}}) \theta$}
        \State{$\theta_{\mathrm{samp}} \leftarrow \eta_{\mathrm{samp}}\theta_{\mathrm{samp}} + (1-\eta_{\mathrm{samp}}) \theta$}
    \Until{\textit{convergence}}
    \State{\textbf{Return} Reward tilted model ${\theta}$}
    \end{algorithmic}
    \end{center}
    \end{minipage}
    }
\end{algorithm}
}

\subsection{Reward-Adaptive CFG based KL Regularization}
\label{sec:cfg_ref_kl}

\paragraph{Motivation: reward hacking under online distribution drift.}
During online fine-tuning, the old model $p_{\mathrm{old}}$ evolves across iterations as $p_\theta$ is updated, inducing distribution drift away from the pretrained model.
This distribution drift increases the risk of \emph{reward hacking} (see \cref{app:online_tilting_reward_hacking} for more discussion): the model exploits weaknesses of the reward model by moving into
out-of-distribution regions, leading to degraded sample quality despite high reward scores.

\paragraph{Mitigation: KL regularization to a fixed initial reference.}
A standard stabilization is to add a KL penalty that anchors the model $\theta$ to a \emph{fixed} initial reference model
(the pretrained model $\phi$), preventing uncontrolled drift \cite{schulman2017proximal,ziegler2019fine,ouyang2022training}:
\begin{equation}
\label{eq:kl_ref_generic}
\beta_{{\mathrm{init}}}\mathrm{KL}\!\left(p_\theta(\cdot\mid c)\;\|\;p_{\phi}(\cdot\mid c)\right)
\!\approx\!
\beta_{\mathrm{init}}\mathbb{E}_{t,\epsilon}\!\left[\lambda(t)\,
\left\|v_\theta(x_t,t\mid c)-v_{\phi}(x_t,t\mid c)\right\|_2^2\right],
\end{equation}
{where $\beta_{\mathrm{init}}$ denote the strength of the initial KL regularization, $\lambda(t)$ is a time-dependent weight which we set to 1 for simplicity, and $v_\phi$ is the initial model velocity prediction.}
However, {in the standard setup of forward-process based methods}, the current model $\theta$ is sampled \emph{without} CFG, while the conditional pretrained model (without CFG) can be significantly weaker than its CFG-guided counterpart.
As a result, anchoring too strongly to a weak non-CFG reference can introduce a training mismatch:
the KL term may dominate and pull the model toward low-quality regions of the pretrained conditional model, meanwhile counteracting reward optimization.

To align the anchoring distribution with the effective sampling semantics of the pretrained model, we instead use a KL penalty with the CFG-guided version of the pretrained model as the fixed reference:
\begin{equation}
\label{eq:kl_cfg_ref}
\beta_{\mathrm{init}}\mathrm{KL}\!\left(p_\theta(\cdot\mid c)\;\|\;p_{\phi}^{\mathrm{CFG}}(\cdot\mid c)\right),
\end{equation}
where $p_{\phi}^{\mathrm{CFG}}(\cdot\mid c)$ denotes the distribution induced by sampling the pretrained model $\phi$ with CFG (guidance scale $s$). 
{This KL term is approximated analogously to \cref{eq:kl_ref_generic}, replacing $v_{\phi}$ with the CFG-guided velocity $v_{\phi}^{\mathrm{CFG}}$.}
{For notation consistency, we rename the \method{} coefficient $\beta$ as $\beta_{\mathrm{old}}$ for moving reference.}

Intuitively, this anchors training to a stronger, more reliable pretrained behavior, which reduces reward hacking while avoiding the
degenerate pull toward the weak non-CFG conditional baseline.
Note that when the RL gradient is missing, this initial KL regularization alone {is functionally equivalent to} CFG distillation \cite{meng2023distillation,chen2024toward,chen2025visual,cideron2024diversity}.

\paragraph{Reward-adaptive initial KL strength for faster optimization.}
While a large initial-reference KL coefficient improves stability, it can also slow learning by excessively restricting policy updates.
Since our raw reward $r_{\mathrm{raw}}(c,x)$ is a scalar and can be normalized to $[0,1]$, we use a simple reward-adaptive scaling of the initial-reference KL strength:
\begin{equation}
\label{eq:beta_ref_adaptive}
\hat{\beta}_{\mathrm{init}}(c,x) \;=\; r_{\mathrm{raw}}(c,x) \, \beta_{\mathrm{init}},
\qquad r_{\mathrm{raw}}(c,x)\in[0,1].
\end{equation}
We apply $\hat{\beta}_{\mathrm{init}}(c,x)$ to the KL term in \eqref{eq:kl_cfg_ref}.
This choice preserves nonnegativity, and it makes the anchoring weaker on low-reward samples (allowing larger corrective updates)
and stronger on high-reward samples (preventing late-stage drift into reward-model loopholes),
empirically accelerating training without sacrificing final performance.
{This strategy coincidences with the take away in GARDO \cite{he2025gardo}.}
Since the adaptive KL strength is sample-dependent, we can finally write our KL loss as:
\begin{equation}
\label{eq:KL_loss}
\mathcal{L}_{\mathrm{KL}} = \frac{1}{K}\sum_{i=1}^K \hat\beta_{\mathrm{init}}(c,x^i)\mathbb{E}_{t,\epsilon}\!\left[\lambda(t)\,
\left\|v_\theta(x_t^i,t\mid c)-v_{\phi}^{\mathrm{CFG}}(x_t^i,t\mid c)\right\|_2^2\right].
\end{equation}

\paragraph{The overall training objective.}
The model $v_\theta$ is trained to minimize the combination of the aforementioned \method{} loss and initial CFG KL regularization:
\begin{equation}
\label{eq:final_loss}
\mathcal{L}_{\mathrm{\method{}\_KL}} = \mathcal{L}_{\mathrm{\method{}}} + \mathcal{L}_{\mathrm{KL}}.
\end{equation}

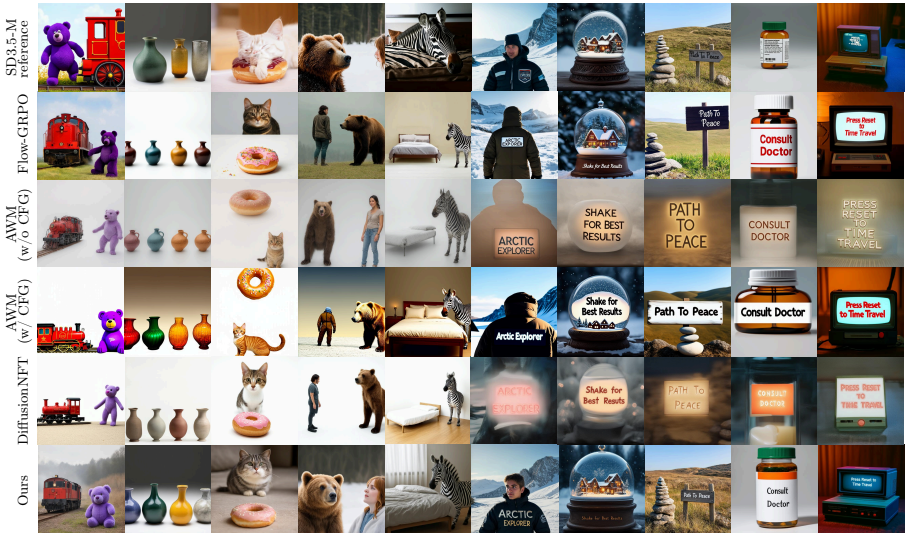
\begin{figure}[t!]
\centering
\begin{overpic}[width=0.98\linewidth]{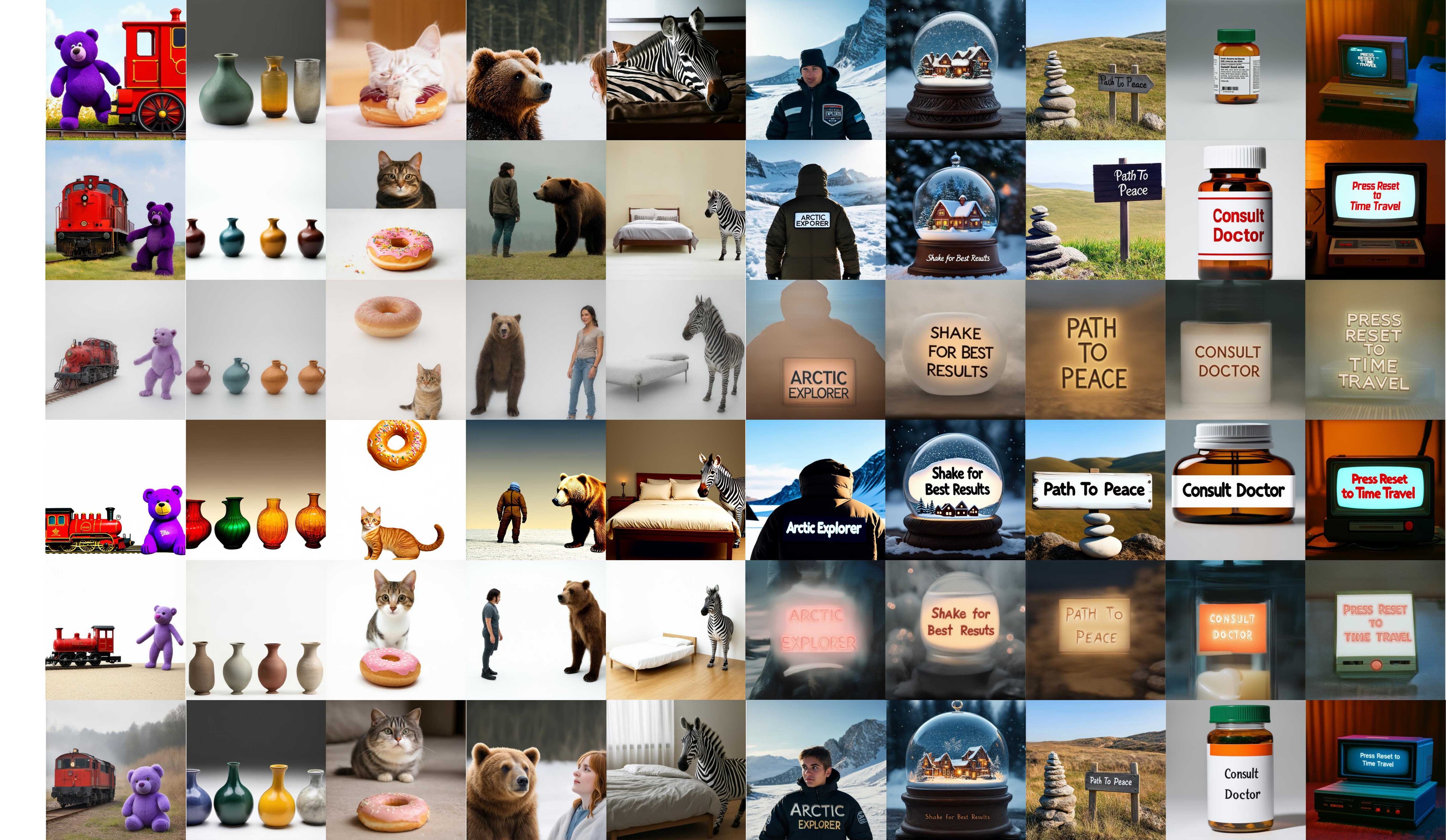}%
\put(-0.4,49.6){\rotatebox{90}{\color{black}\scalebox{0.6}{ SD3.5-M}}}
\put(0.9,49.4){\rotatebox{90}{\color{black}\scalebox{0.6}{ reference}}}
\put(0.9,38.8){\rotatebox{90}{\color{black}\scalebox{0.6}{ Flow-GRPO}}}
\put(-0.4,31.2){\rotatebox{90}{\color{black}\scalebox{0.6}{ AWM }}}
\put(0.9,29.2){\rotatebox{90}{\color{black}\scalebox{0.6}{ (w/o CFG)}}}
\put(-0.4,21.5){\rotatebox{90}{\color{black}\scalebox{0.6}{ AWM}}}
\put(0.9,20.1){\rotatebox{90}{\color{black}\scalebox{0.6}{ (w/ CFG)}}}
\put(0.9,9.){\rotatebox{90}{\color{black}\scalebox{0.6}{ DiffusionNFT}}}
\put(0.9,2.3){\rotatebox{90}{\color{black}\scalebox{0.6}{ Ours}}}
\end{overpic}
\caption{
\small Visual comparison between benchmarks and our models. 
}
\label{fig:visual_comparison}
\vspace{-15pt}
\end{figure}

\section{Experiments}
\label{sec:experiments}

\begin{table*}[t]
    \centering
    \caption{\textbf{Performance on Compositional Image Generation, Visual Text Rendering, and Human Preference benchmarks.} Task metrics are evaluated on the corresponding test prompts; image quality and preference scores are evaluated on \texttt{DrawBench} \cite{saharia2022photorealistic} prompts. Baseline results are taken from Flow-GRPO~\cite{liu2025flow} or reimplemented when unavailable in original papers. \texttt{ImgRwd}: \texttt{ImageReward}.}
    \resizebox{\linewidth}{!}{
        \begin{tabular}{lccccccc}
            \toprule
            \multirow{2}{*}{\textbf{Model}} & \multicolumn{2}{c}{\textbf{Task Metric}} & \textbf{Image Quality} & \multicolumn{4}{c}{\textbf{Preference Score}}    \\ \cmidrule(lr){2-3} \cmidrule(lr){4-4} \cmidrule(l){5-8} 
                                   & \textbf{\texttt{GenEval}} $\uparrow$  & \textbf{\texttt{OCR}} $\uparrow$ & \textbf{\texttt{Aesthetics}} $\uparrow$& \textbf{\texttt{ImgRwd}} $\uparrow$& \textbf{\texttt{PickScore}} $\uparrow$ & \textbf{\texttt{CLIPScore}} $\uparrow$& \textbf{\texttt{HPSv2.1}} $\uparrow$ \\ \midrule
            SD3.5-M & 0.63 & 0.59 & 5.39 & 0.87 & 22.34 & 27.99 & 0.279 \\
            \midrule
            \multicolumn{8}{c}{\textit{Compositional Image Generation}} \\
            \midrule
            Flow-GRPO~\cite{liu2025flow} (w/o KL) & 0.95 & — & 4.93 & 0.44 & 21.16 & — & — \\
            Flow-GRPO~\cite{liu2025flow} (w/ KL)  & 0.95 & — & 5.25 & 1.03 & 22.37 & 29.25 & 0.274 \\
            AWM~\cite{xue2025advantage} (w/o CFG) & 0.91 & — & 5.10 & 0.39 & 21.76 & 26.99 & 0.235 \\
            AWM~\cite{xue2025advantage} (w/ CFG) & 0.86 & — & 5.25 & 1.06 & 22.12 & 28.86 & 0.278 \\
            DiffusionNFT~\cite{zheng2025diffusionnft} & 0.92 & — & 5.30 & 0.63 & 22.07 & 27.25 & 0.253 \\
            \rowcolor{gray!20} \method{} & 0.93 & — & 5.44 & 0.98 & 22.48 & 28.44 & 0.284 \\
            \rowcolor{gray!20} \qquad + CFG sampling=1.5 & 0.93 & — & 5.42 & 1.05 & 22.55 & 28.80 & 0.289 \\
            \rowcolor{gray!20} \qquad + CFG sampling=3.0 & 0.92 & — & 5.37 & 1.10 & 22.43 & 29.00 & 0.287 \\
            
            \midrule
            \multicolumn{8}{c}{\textit{Visual Text Rendering}} \\
            \midrule
            Flow-GRPO~\cite{liu2025flow} (w/o KL) & — & 0.93 & 5.13 & 0.58 & 21.79 & — & — \\
            Flow-GRPO~\cite{liu2025flow} (w/ KL)  & — & 0.92 & 5.32 & 0.95 & 22.44 & 28.86 & 0.282 \\
            AWM~\cite{xue2025advantage} (w/o CFG) & — & 0.97 & 5.22 & -0.64 & 20.82 & 24.54 & 0.203 \\
            AWM~\cite{xue2025advantage} (w/ CFG) & — & 0.96 & 5.35 & 0.96 & 22.36 & 28.70 & 0.283 \\
            DiffusionNFT~\cite{zheng2025diffusionnft} & — & 0.97 & 4.89 & -0.81 & 20.53 & 24.08 & 0.196 \\
            \rowcolor{gray!20} \method{} & — & 0.92 & 5.33 & 0.87 & 22.40 & 28.48 & 0.281 \\
            \rowcolor{gray!20} \qquad + CFG sampling=1.5 & — & 0.92 & 5.31 & 0.97 & 22.50 & 28.77 & 0.287 \\
            \rowcolor{gray!20} \qquad + CFG sampling=3.0 & — & 0.87 & 5.28 & 1.04 & 22.44 & 29.16 & 0.290 \\
            \bottomrule
            \end{tabular}
}
\vspace{-15pt}
\label{tab:main}
\end{table*}

\subsection{Experiment Setup}

\paragraph{Models and training configuration.}
All experiments fine-tune Stable Diffusion 3.5-Medium (SD3.5-M) \cite{esser2024scaling} and following Flow-GRPO \cite{liu2025flow}. 
Unless stated otherwise, images are generated at $512\times512$ resolution with a group size $K=24$ for the main results and a group size $K=6$ for all the ablations.
We train LoRA adapters \cite{hu2022lora} with $r=32$ and $\alpha=64$, and adopt the remaining optimization and implementation details from DiffusionNFT \cite{zheng2025diffusionnft}. 
All main experiments are run on up to four NVIDIA H100 GPUs and ablations on a single NVIDIA H100 GPU, each completing in under 36 hours.

\paragraph{Datasets and reward models.}
We study two non-differentiable reward settings: (i) compositional image generation using \texttt{GenEval} \cite{ghosh2023geneval} as the reward, and (ii) visual text rendering using an \texttt{OCR} reward \cite{chen2023textdiffuser}. For both tasks, we use the same training/evaluation prompt splits and the corresponding reward models as in Flow-GRPO \cite{liu2025flow}.
To assess generalization (reward hacking) and overall image quality beyond the task reward, we additionally evaluate on \texttt{DrawBench} \cite{saharia2022photorealistic} prompts and report a suite of metrics, including \texttt{PickScore} \cite{kirstain2023pick}, \texttt{CLIPScore} \cite{hessel2021clipscore}, \texttt{HPSv2.1} \cite{wu2023human}, \texttt{Aesthetics} \cite{schuhmann2022aesthetics}, and \texttt{ImageReward} \cite{xu2023imagereward}, capturing image quality, image-text alignment, and human preference.

Further experimental details are provided in \cref{sec:expr_setup}.

\subsection{Main Results}

\begin{table*}[t]
    \centering
    \caption{{\bf \texttt{GenEval} Result.} 
    Best scores are in \colorbox{blue!10}{blue}, second-best in \colorbox{green!15}{green}.
    Results for baselines are taken from Flow-GRPO~\cite{liu2025flow} or the respective original papers, and reimplemented where unavailable. Obj.: Object; Attr.: Attribution.
    }
    \vspace{-5pt}
    \resizebox{\linewidth}{!}{
    \begin{tabular}{lccccccc}
    \toprule
    \textbf{Model} & \textbf{Overall} & \textbf{Single Obj.} & \textbf{Two Obj.} & \textbf{Counting} & \textbf{Colors} & \textbf{Position} & \textbf{Attr. Binding} \\
    \cmidrule(lr){2-2} \cmidrule(lr){3-8}
    \midrule
    \multicolumn{8}{c}{\textit{Autoregressive Models}} \\
    \midrule
    Show-o~\cite{xie2024show}  & 0.53 & 0.95 & 0.52 & 0.49 & 0.82 & 0.11 & 0.28 \\
    Janus-Pro-7B~\cite{chen2025janus}  & 0.80 & \colorbox{green!15}{0.99} & 0.89 & 0.59 & \colorbox{green!15}{0.90} & {0.79} & {0.66} \\
    GPT-4o~\cite{gpt4o} & {0.84} & \colorbox{green!15}{0.99} & 0.92 & 0.85 & \colorbox{blue!10}{0.92} & 0.75 & 0.61 \\
    \midrule
    \multicolumn{8}{c}{\textit{Diffusion and Flow Matching Models}} \\
    \midrule
    SD-XL~\cite{podell2023sdxl}  & 0.55 & 0.98 & 0.74 & 0.39 & 0.85 & 0.15 & 0.23 \\
    FLUX.1 Dev~\cite{flux2024}  & 0.66 & 0.98 & 0.81 & 0.74 & 0.79 & 0.22 & 0.45 \\
    SD3.5-M~\cite{esser2024scaling} & 0.63 & 0.98 & 0.78 & 0.50 & 0.81 & 0.24 & 0.52 \\
    \midrule
    \multicolumn{8}{c}{\textit{Flow + RL}} \\
    \midrule
    {SD3.5-M+Flow-GRPO~\cite{liu2025flow}} & \colorbox{blue!10}{0.95} & \colorbox{blue!10}{1.00} & \colorbox{blue!10}{0.99} & \colorbox{green!15}{0.95} & \colorbox{blue!10}{0.92} & \colorbox{blue!10}{0.99} & \colorbox{blue!10}{0.86}  \\
    {SD3.5-M+AWM~\cite{xue2025advantage}} (w/o CFG) & 0.91 & \colorbox{blue!10}{1.00} & \colorbox{green!15}{0.98} & \colorbox{green!15}{0.95} & 0.77 & 0.78 & 0.65  \\
    {SD3.5-M+DiffusionNFT~\cite{zheng2025diffusionnft}} & 0.92 & \colorbox{blue!10}{1.00} & \colorbox{blue!10}{0.99} & \colorbox{blue!10}{0.96} & 0.82 & 0.78 & \colorbox{green!15}{0.73}  \\
    \midrule
    {SD3.5-M+\method{}} & \colorbox{green!15}{0.93} & \colorbox{blue!10}{1.00} & \colorbox{green!15}{0.98} & 0.92 & 0.88 & \colorbox{green!15}{0.90} &  \colorbox{green!15}{0.73} \\
    \bottomrule
    \end{tabular}
    }
\label{tab:geneval}
\end{table*}

\paragraph{Compositional image generation.}
In \cref{tab:main}, CRD achieves a \texttt{GenEval} score of 0.93 (vs.\ 0.63 for SD3.5-M), and offers a strong quality--alignment trade-off. While Flow-GRPO \cite{liu2025flow} reaches the highest \texttt{GenEval} (0.95), CRD attains the best \texttt{Aesthetics} (5.44) and strong preference metrics (\texttt{ImageReward}: 0.98, \texttt{PickScore}: 22.48, \texttt{HPSv2.1}: 0.284). Notably, CRD is the only method that improves \texttt{Aesthetics} over SD3.5-M (5.44 vs.\ 5.39) while also increasing preference scores.
The detailed \texttt{GenEval} performance comparison is provided in \cref{tab:geneval}.

\paragraph{Visual text rendering.}
CRD obtains 0.92 \texttt{OCR} accuracy, matching Flow-GRPO (w/ KL), while maintaining positive preference metrics on unseen prompts ( \texttt{ImageReward}: 0.87, \texttt{PickScore}: 22.40, \texttt{HPSv2.1}: 0.281). In contrast, DiffusionNFT~\cite{zheng2025diffusionnft} and AWM~\cite{xue2025advantage} (w/o CFG) incur negative \texttt{ImageReward} ($-$0.81/$-$0.64), indicating degraded perceptual quality despite high \texttt{OCR} score.

\paragraph{Qualitative comparison.}
\Cref{fig:visual_comparison} shows CRD preserves SD3.5-M photorealism while improving prompt fidelity, especially for multi-object composition and integrated text. Flow-GRPO often yields less natural backgrounds, AWM (w/ CFG) over-saturates toward a stylized look, and DiffusionNFT tends to produce uniform backgrounds or blurred regions around text. CRD produces more legible, naturally embedded text and coherent object relations, consistent with \cref{tab:main}.

\paragraph{{Global} discussion on results.}
As shown in \cref{tab:main,fig:visual_comparison}, although AWM does not require CFG during training, this setting leads to reward hacking and degraded visual quality; incorporating CFG partially alleviates this but introduces over-saturation artifacts.
Although CRD implicitly distills the CFG guidance through \cref{eq:KL_loss}, moderate test-time CFG further improves preference scores (\cref{tab:main}), likely by increasing effective text-conditioning strength and due to the insufficient `CFG distillation' in our RL fine-tuning.
Based on the qualitative results in \cref{fig:visual_comparison}, we also notice that images generated with our method are more close to the reference model than the {ones from the} competing methods.
Finally, although \method{} underperforms Flow-GRPO on certain metrics, it inherits the efficiency of forward-process-based training and converges significantly faster, offering a more practical trade-off.

Additional experimental results can be found in \cref{sec:app_additional_result}.

\subsection{Ablation Studies}
\label{sec:ablation}

We study the impact of three core design choices: the old-model decay rate $\eta_{\mathrm{old}}$, the initial KL strength $\beta_{\mathrm{init}}$, and the adaptive KL schedule $\hat{\beta}_{\mathrm{init}}$.
More ablations can be found in \cref{sec:app_additional_result}.

\begin{figure*}[t!]
    \centering
    \begin{subfigure}[t]{0.325\textwidth}
        \centering
        \includegraphics[width=\textwidth]{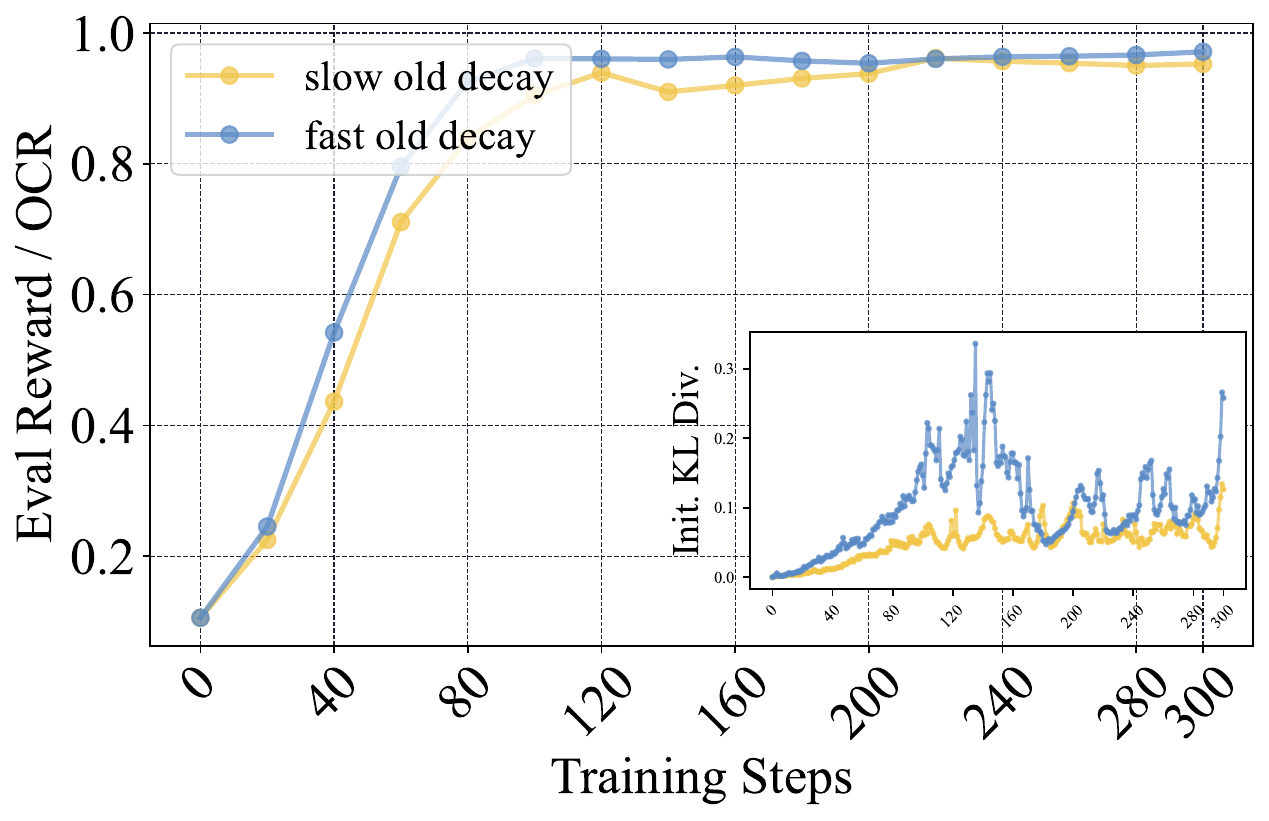}
        \subcaption{\resizebox{0.85\linewidth}{!}{\small Slow and fast old decay $\eta_{\mathrm{old}}$}
        }
        \label{subfig:ablation1}
    \end{subfigure}
    \hfill
    \begin{subfigure}[t]{0.325\textwidth}
        \centering
        \includegraphics[width=\textwidth]{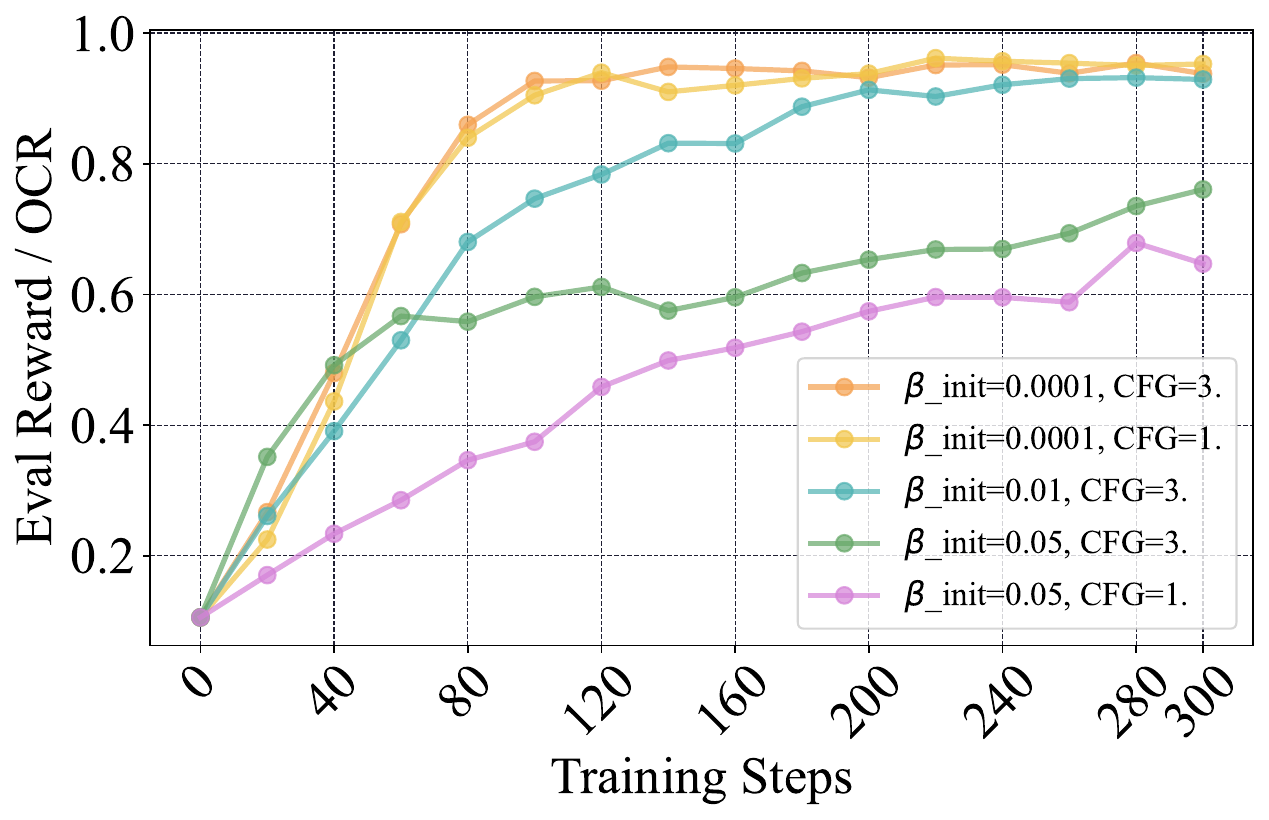}
        \subcaption{\resizebox{0.85\linewidth}{!}{\small Results with different $\beta_{\mathrm{init}}$}
        }
        \label{subfig:ablation2}
    \end{subfigure}
    \hfill
    \begin{subfigure}[t]{0.325\textwidth}
        \centering
         \includegraphics[width=\textwidth]{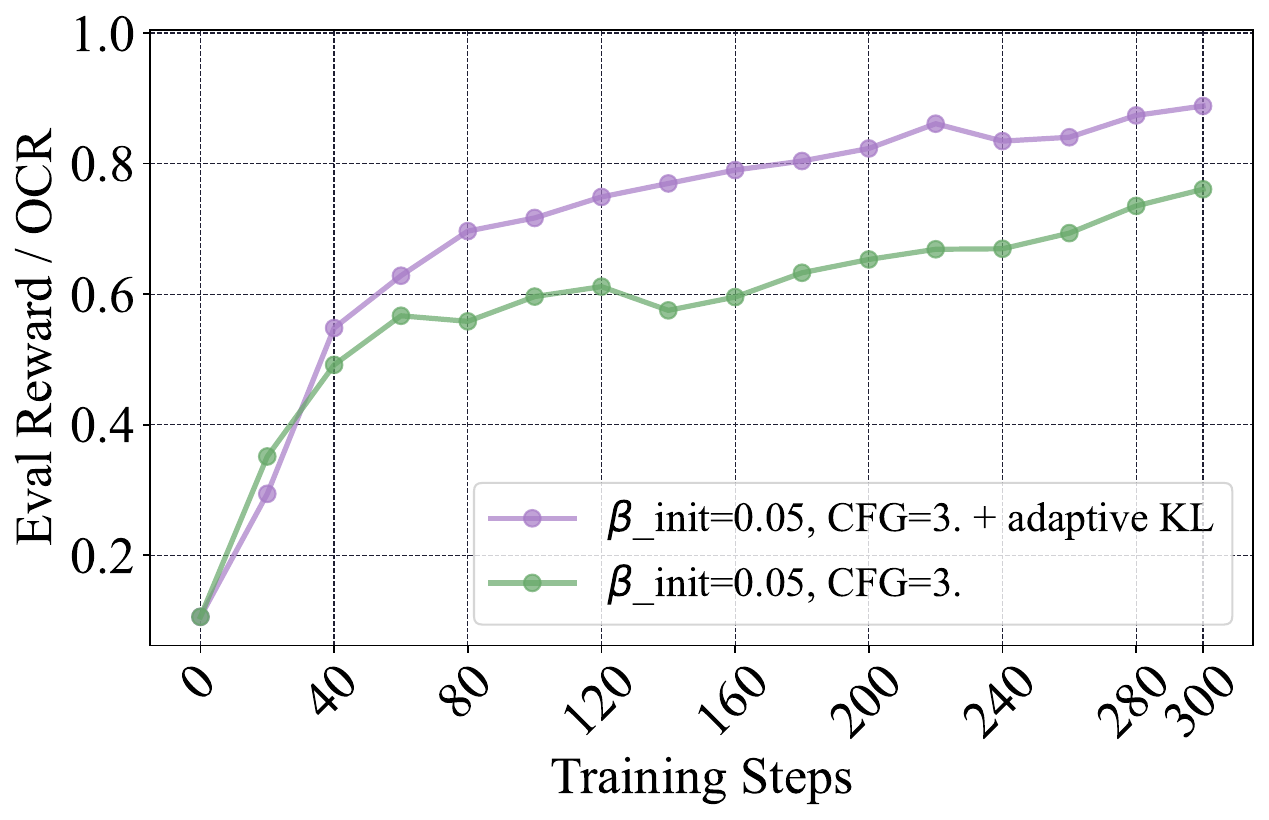}
        \vspace{-13.5pt}
        \subcaption{\resizebox{0.85\linewidth}{!}{\small Ablation on adaptive $\hat{\beta}_{\mathrm{init}}$}}
        \label{subfig:ablation3}
    \end{subfigure}
    \vspace{-8pt}
    \caption{
    \small
    Ablations on slow old model decay rate $\eta_{\mathrm{old}}$, and initial KL strength $\beta_{\mathrm{init}}$.}
    \label{fig:Ablation}
    \vspace{10pt}
\end{figure*}

\begin{figure}[t!]
\centering
\begin{overpic}[width=0.98\linewidth]{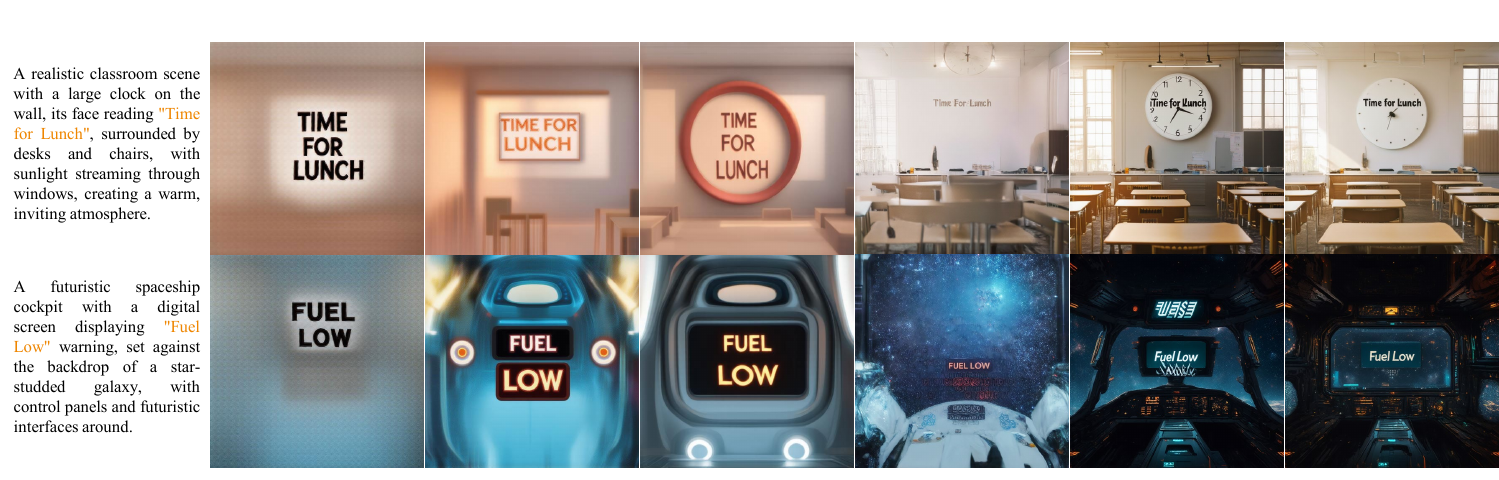}%
\put(15.5,29.9){\color{black}\scalebox{0.65}{ fast old decay}}
\put(29.9,29.9){\color{black}\scalebox{0.65}{ $\beta_{\mathrm{init}}$=0.0001}}
\put(29.9,31.9){\color{black}\scalebox{0.65}{ CFG=1.0}}
\put(29.9,33.9){\color{black}\scalebox{0.65}{ slow old decay}}
\put(45.,29.9){\color{black}\scalebox{0.65}{ +CFG=3.0}}
\put(59.65,29.9){\color{black}\scalebox{0.65}{ $\beta_{\mathrm{init}}$=0.05}}
\put(59.65,31.9){\color{black}\scalebox{0.65}{ CFG=1.0}}
\put(73.65,29.9){\color{black}\scalebox{0.65}{ +CFG=3.0}}
\put(86.15,29.9){\color{black}\scalebox{0.65}{ +Adaptive $\hat{\beta}_{\mathrm{init}}$}}
\end{overpic}
\caption{
\small Visual comparison corresponding to the ablations in \cref{fig:Ablation}.
}
\label{fig:visual_ablation}
\vspace{-15pt}
\end{figure}

\paragraph{Old-model decay (slow vs.\ fast).}
We first validate the proposed \emph{slow} old-model decay. As shown in \cref{subfig:ablation1}, using a \emph{fast} decay yields faster convergence, but it also causes highly unstable KL divergence with model $\phi$. This instability and large KL value usually correspond to severe reward hacking, consistent with the qualitative results in \cref{fig:visual_ablation}{: the fast-decay generations collapse to plain text on flat backgrounds, satisfying the \texttt{OCR} reward while losing all scene coherence.}

\paragraph{Initial KL strength.}
In \cref{subfig:ablation2}, we report evaluation reward curves under different initial KL coefficients $\beta_{\mathrm{init}}$ and CFG values. We observe a clear trade-off between the strength of the initial KL constraint and training speed, where smaller $\beta_{\mathrm{init}}$ accelerates optimization. Moreover, the visual comparisons in \cref{fig:visual_ablation} indicate that a larger initial KL typically produces better perceptual quality.

\paragraph{Adaptive KL.}
In \cref{subfig:ablation3}, we evaluate {the proposed} adaptive weighting for the initial KL coefficient. The adaptive strategy improves training progress while incurring minimal degradation in generation quality, as supported by both the reward curves and the qualitative results in \cref{fig:visual_ablation}.

\section{Conclusion}
\label{sec:conclusion}

We presented \textbf{Centered Reward Distillation (\method{})}, a forward-process diffusion RL framework grounded in KL-regularized reward maximization. 
Our key insight, that the intractable prompt-dependent normalizer cancels under within-prompt reward centering, yields a well-posed family of reward-matching objectives that unifies prior methods as special cases. 
Paired with practical techniques targeting distribution drift and reward hacking, \method{} achieves competitive reward optimization with fast and stable training on \texttt{GenEval} and \texttt{OCR} 
benchmarks, and competitive results on unseen metrics.

\clearpage

\bibliographystyle{splncs04}
\bibliography{main}

\clearpage
\appendix
\setcounter{page}{1}

\begin{center}
    \Large\textbf{
    Appendix for \method{}}
\end{center}

This appendix is organized as follows:
\begin{itemize}
    \item \cref{sec:limitations}: Limitation and future work.
    \item \cref{supp:discussion}: More discussion on related works.
    \item \cref{sec:expr_setup}: Experimental details.
    \item \cref{app:diffusionnft_kl_equivalence}: Theoretical discussion on DiffusionNFT.
    \item \cref{app:online_tilting_reward_hacking}: Discussion of accumulated tilting.
    \item \cref{supp:special_case}: GVPO and Reward Distill as special cases of \method{}.
    \item \cref{supp:ratio_based}: Ratio-based Reward Distillation and connection to InfoNCA.
    \item \cref{sec:app_additional_result}: Additional experimental results and visualizations.
\end{itemize}

\vspace{0.2cm}

\section{Limitations and Future Work}
\label{sec:limitations}
Despite strong empirical performance, this work has several limitations that suggest clear directions for future research. 
\textit{First}, while \method{} is motivated by a KL-regularized optimality view, a more rigorous theoretical analysis remains open including convergence and stability guarantees under the practical approximations such as ELBO-based log-ratio surrogates, moving references, and finite-sample within-prompt centering.
\textit{Second}, \method{} is fundamentally constrained by the coverage and capabilities of the {reference} (pretrained) model, since training relies on self-generated samples: if the base model rarely produces trajectories that exhibit a desired behavior, the reward signal provides little leverage to bootstrap it, and continued self-training can further narrow diversity. A natural mitigation is to inject real data and explicit supervision into the training pipeline \cite{chen2025bridging}. 
For example, mixing RL updates with Supervised Fine-Tuning (SFT) updates or adding a data-anchoring regularizer (data KL) to preserve coverage and diversity \cite{ye2025data}.  
\textit{Finally}, although \method{} is substantially more efficient than Flow-GRPO variants, it still incurs non-trivial computational overhead relative to standard fine-tuning as it requires full sampling during training. This cost becomes more pronounced for video diffusion models with long temporal trajectories and high-resolution frames \cite{wan2025wan}. Future work could reduce training costs through integration with faster samplers or diffusion distillation techniques to amortize generation costs while maintaining reward optimization performance.

\section{More Discussion with Related Works}
\label{supp:discussion}

The objective of prior work GVPO \cite{zhang2025gvpo} can be viewed as a special case of our \method{}. GVPO focuses on the variance analysis of the loss gradient and is primarily designed for LLM post-training. While the two methods share conceptual similarities, \method{} is derived from a different perspective and is tailored for image generation tasks with text-to-image diffusion models. 
Furthermore, we demonstrate that \method{} provides a unified framework that subsumes both GVPO and prior two-sample reward distillation methods \cite{mao2024don,gao2024rebel,fisch2024robust}.

{More recently, Ou \etal \cite{ou2026diffusion} introduce Variance Minimisation Policy Optimisation (VMPO), which formulates diffusion alignment as minimising the variance of log importance weights rather than directly optimising a KL-based objective. Despite the surface-level similarity in the final objective, VMPO is more closely related to a diffusion-adapted variant of GVPO \cite{zhang2025gvpo}, accompanied by corresponding theoretical analysis. 
Compared to \method{}, VMPO adopts a different empirical focus, and at the time of our submission, its scope remains narrower in terms of the experimental coverage that \method{} provides.}

Finally, for diffusion RL, all forward-process-based methods address different aspects of a broader picture. Together, they form a complementary recipe that enables faster training while mitigating reward hacking.

\section{Experimental Details}
\label{sec:expr_setup}

\method{} is implemented on top of the DiffusionNFT codebase, with hyperparameters and model configurations listed in \cref{table:hyperparameters}.
Regarding training time, we compare with AWM and DiffusionNFT under the same training budget; however, as both baselines suffer from reward hacking, we report their results at early checkpoints where this issue is less severe.
Throughout the experiments, we adopt the adaptive weighting~\cite{yin2024one,zheng2025diffusionnft} variant of the implicit model reward:
\begin{equation}
\label{eq:rwrd_policy_reward_elbo_adaptive}
\widehat{R}_\theta^{\mathrm{adap}}(c,x)
\;\triangleq\;
-\beta\,
\mathbb{E}_{t,\epsilon}\!\left[
\frac{d\left\|v_\theta(x_t,t\mid c)-v_\mathrm{target}\right\|_2^2}{\mathrm{sg}(\left\|v_\theta(x_t,t\mid c)-v_\mathrm{target}\right\|_1)}
-
\frac{d\left\|v_{\mathrm{ref}}(x_t,t\mid c)-v_\mathrm{target}\right\|_2^2}{\mathrm{sg}(\left\|v_{\mathrm{ref}}(x_t,t\mid c)-v_\mathrm{target}\right\|_1)}
\right],
\end{equation}
where $d$ is the dimension of $x$ and $\mathrm{sg}(\cdot)$ denotes the stop-gradient operation.

\begin{table}[t!]
  \centering
  \setlength{\tabcolsep}{5pt}
  \renewcommand{\arraystretch}{1.2}
  \vspace{-0.3cm}
  \caption{
    \small
    Hyperparameters {and configurations} used for main experiment runs. Model architecture is the same as the pretrained model. 
    Largest Experiments in this work are conducted with 4 NVIDIA H100 GPUs.
    $i$ in $\eta_{\mathrm{old}}$ and $\eta_{\mathrm{init}}$ denotes for \textit{Training (Optimizer) Steps}.
    }
  \resizebox{0.98\linewidth}{!}{%
  \begin{tabular}{|l|l|c|c|c|}
    \hline
    & Reward Model & \texttt{OCR} & \texttt{GenEval} & \texttt{Multi Rewards} \\
    \hline
    \multirow{2}{*}{LoRA}
    & $\alpha$  & 64 & 64 & 64  \\
    & $r$ & 32 & 32 & 32  \\
    \hline
    \multirow{4}{*}{Sampling}
    & Sampling CFG  & 1.0 & 1.0 & 1.0  \\
    & Training Rollout Steps & 10 & 10 & 15  \\
    & Evaluation Sampling Steps & 40 & 40 & 40  \\
    & ODE Solver & `dpmv2' & `dpmv2' & `dpmv2'  \\
    \hline
    \multirow{5}{*}{Forward-Process RL}
    & {Generation Image Resolution} & 512$\times$512 & 512$\times$512 & 512$\times$512 \\
    & Group Size $K$ & 24 & 24 & 24  \\
    & Training (Optimizer) Steps & 720 & 360 & 300 \\
    & Num of Groups per Batch & 48 & 48 & 48 \\
    & Optimizer Step per Batch & 2 & 1 & 1 \\
    \hline
    \multirow{6}{*}{\method{}}
    & $\beta_{\mathrm{old}}$ & 1.0 & 1.0 & 0.1 \\
    & $\beta_{\mathrm{init}}$ & 0.1 & 0.08 & 0.01 \\
    & CFG for Initial KL & 4.5 & 4.5 & 4.5 \\
    & Use Adaptive KL Strength $\hat{\beta}_{\mathrm{init}}$ & True & True & True \\
    & $\eta_{\mathrm{old}}$ & $\min(0.25+0.005i,\; 0.999)$ & $\min(0.1+0.001i,\;0.5)$ & $\min(0.5+0.0025i,\; 0.999)$ \\
    & $\eta_{\mathrm{init}}$ & $\min(\max(0.0075(i-75),\;0),\;0.999)$ & $\min(0.001i,\;0.5)$ & $\min(0.001i,\;0.5)$ \\
    \hline
    \multirow{5}{*}{Optimization}
    & Optimizer & AdamW & AdamW & AdamW  \\
    & $(\beta_1,\beta_2)$ & (0.9,0.999) & (0.9,0.999) & (0.9,0.999) \\
    & Learning Rate & 3e-4 & 3e-4 & 3e-4  \\
    & Weight Decay & 1e-4 & 1e-4 & 1e-4  \\
    & EMA Ratio & 0.9 & 0.9 & 0.9 \\
    \hline
  \end{tabular}}
  \vspace{-0.3cm}
\label{table:hyperparameters}
\end{table}

\section{DiffusionNFT Recovers the KL-Regularized Exponential-Tilt Optimum}
\label{app:diffusionnft_kl_equivalence}

{Unlike other formulations that explicitly include the log density ratio between the current and reference models and rely on a diffusion ELBO estimator to approximate it (an approach that generalizes naturally to LLMs), DiffusionNFT is derived directly for diffusion and flow models without invoking the log density ratio, making it a `native' diffusion RL algorithm.
Nevertheless, we show in this section that when the reference model is held fixed, DiffusionNFT is 
equivalent to optimizing the same \emph{exponentially tilted} target distribution as standard KL-regularized reward maximization (see \cref{eq:rwrd_boltzmann}), revealing a deep connection between DiffusionNFT and our method.
We present the argument at the density level using the notation of the main text.}

\subsection{DiffusionNFT Score Update Implies an Exponential Tilt}
DiffusionNFT is defined through a modification of the velocity (or score) field of the diffusion model relative to the
old model. Let $p^{*}(\cdot\mid c)$ denote the population-optimal distribution induced by the optimal velocity field
$v^{*}$ for a \emph{fixed} old/reference model $p_{\mathrm{old}}$.
Under the fixed Gaussian noising family, {for two diffusion processes $A$ and $B$,} the differences of velocity fields reduce to differences of score functions:
\begin{equation}
\label{eq:app_v_to_score}
v^{A}(x_t,t\mid c)-v^{B}(x_t,t\mid c)
\;=\;
\kappa(t)\Big(\nabla_{x_t}\log p^{A}_t(x_t\mid c)-\nabla_{x_t}\log p^{B}_t(x_t\mid c)\Big),
\end{equation}
where $\kappa(t)$ depends only on the noise schedule, and {$p^A_t(\cdot\mid c)$ and $p^B_t(\cdot\mid c)$ denote the noised marginals at time $t$
induced by diffusion process $A$ and $B$, respectively. }

DiffusionNFT constructs a guidance direction of the form
\begin{equation}
\label{eq:app_delta_def}
\Delta(x_t,c,t)
\;=\;
\alpha(x_t,c)\bigl(v^{+}(x_t,t\mid c)-v^{\mathrm{old}}(x_t,t\mid c)\bigr),
\end{equation}
where $\alpha(x_t,c)$ is an ``optimality'' score (in DiffusionNFT, an optimality posterior).
Using the Bayes relation
\begin{equation}
\label{eq:app_bayes_positive}
p^{+}_t(x_t\mid c)
\;=\;
p^{\mathrm{old}}_t(x_t\mid c)\,\frac{\alpha(x_t,c)}{\mathbb{E}_{x_t\sim p^{\mathrm{old}}_t(\cdot\mid c)}[\alpha(x_t,c)]},
\end{equation}
and applying \cref{eq:app_v_to_score}, one obtains
\begin{equation}
\label{eq:app_vplus_minus_vold}
v^{+}(x_t,t\mid c)-v^{\mathrm{old}}(x_t,t\mid c)
\;=\;
\kappa(t)\nabla_{x_t}\log\alpha(x_t,c),
\end{equation}
and therefore
\begin{equation}
\label{eq:app_delta_simplified}
\Delta(x_t,c,t)
\;=\;
\kappa(t)\,\alpha(x_t,c)\,\nabla_{x_t}\log\alpha(x_t,c)
\;=\;
\kappa(t)\,\nabla_{x_t}\alpha(x_t,c).
\end{equation}

In the population optimum of DiffusionNFT (cf.\ the optimality condition in the main text), the optimal velocity satisfies {(see Eq. (3) in DiffusionNFT)}
\begin{equation}
\label{eq:app_optimal_v_condition}
v^{*}(x_t,t\mid c)-v^{\mathrm{old}}(x_t,t\mid c)
\;=\;
\frac{2}{\beta}\,\Delta(x_t,c,t).
\end{equation}
Combining \cref{eq:app_v_to_score}, \cref{eq:app_delta_simplified}, and \cref{eq:app_optimal_v_condition} yields a score-level characterization:
\begin{equation}
\label{eq:app_score_ratio_grad}
\nabla_{x_t}\log\frac{p^{*}_t(x_t\mid c)}{p^{\mathrm{old}}_t(x_t\mid c)}
\;=\;
\frac{2}{\beta}\,\nabla_{x_t}\alpha(x_t,c).
\end{equation}
Integrating both sides over $x_t$ gives the corresponding density-level optimum:
\begin{equation}
\label{eq:app_density_tilt}
p^{*}_t(x_t\mid c)
\;=\;
\frac{1}{Z_t(c)}\;p^{\mathrm{old}}_t(x_t\mid c)\;
\exp\!\left(\frac{2}{\beta}\,\alpha(x_t,c)\right),
\end{equation}
where $Z_t(c)=\int p^{\mathrm{old}}_t(x_t\mid c)\exp\!\left(\frac{2}{\beta}\alpha(x_t,c)\right)\,dx_t$.

\subsection{Reduction to KL-regularized RL at $t=0$}
At the data level ($t=0$), we identify $\alpha(x_0,c)$ with the (scaled) reward used for training; in our notation we take
\begin{equation}
\label{eq:app_alpha_to_reward}
\alpha(x_0,c)\equiv r(c,x_0).
\end{equation}
Substituting \cref{eq:app_alpha_to_reward} into \cref{eq:app_density_tilt} at $t=0$ yields
\begin{equation}
\label{eq:app_t0_tilt}
p^{*}(x_0\mid c)
\;\propto\;
p^{\mathrm{old}}(x_0\mid c)\exp\!\left(\frac{2}{\beta}\,r(c,x_0)\right).
\end{equation}
\cref{eq:app_t0_tilt} is exactly the KL-regularized exponential-tilt optimum \cref{eq:rwrd_boltzmann}, with reference
$p_{\mathrm{ref}}=p_{\mathrm{old}}$ and an effective inverse-temperature $\lambda=2/\beta$.
Therefore, \emph{with a fixed reference model}, DiffusionNFT is equivalent (at the population optimum) to KL-regularized reward
maximization targeting the same exponentially tilted distribution (see \cref{eq:rwrd_boltzmann,eq:app_t0_tilt}).

\paragraph{Implication.}
This equivalence makes explicit that DiffusionNFT performs KL-regularized policy improvement relative to $p_{\mathrm{old}}$,
and also clarifies why repeatedly updating the reference online can accumulate tilt (see \cref{app:online_tilting_reward_hacking}).

\section{Online RL Leads to Reward Hacking via Accumulated Tilting}
\label{app:online_tilting_reward_hacking}

This section explains why running KL-regularized diffusion fine-tuning \emph{online} with a moving reference can induce
\emph{accumulated tilting} \cite{peters2007reinforcement,kim2025test},
leading to increasingly peaked policies and potential reward hacking (see \cref{fig:visual_comparison,fig:visual_ablation}).

\paragraph{Online training with a moving reference.}
In practice, methods such as DiffusionNFT can be run online: after optimizing the student model $p_\theta$ for one epoch,
the reference model is updated (e.g., by a hard copy of weights or by an Exponential Moving Average (EMA)),
and the next epoch optimizes against this updated reference.
This induces a recursion over the reference distributions across epochs.

\subsection{Idealized Recursion under Exact Per-epoch Optima}
Let $p^{(k)}(x\mid c)$ denote the reference (``old'') model distribution at epoch $k$.
We idealize the analysis by assuming that (i) the reward function $r(c,x)$ is fixed across epochs and
(ii) at each epoch we reach the population optimum of the KL-regularized objective with inverse-temperature $\beta>0$.
Under this idealization, the population-optimal model at epoch $k+1$ is the exponentially tilted version of the reference:
\begin{equation}
\label{eq:app_online_tilting_recursion}
p^{(k+1)}(x\mid c)
\;=\;
\frac{1}{Z^{(k)}(c)}\; p^{(k)}(x\mid c)\;
\exp\!\left(\frac{1}{\beta}\, r(c,x)\right),
\end{equation}
where $Z^{(k)}(c)=\int p^{(k)}(x\mid c)\exp(r(c,x)/\beta)\,dx$ is the normalizer. \cref{eq:app_online_tilting_recursion} is the same Boltzmann form as in
\cref{eq:rwrd_boltzmann}, applied epoch-by-epoch with the updated reference.

Unrolling the recursion yields
\begin{equation}
\label{eq:app_online_tilting_unrolled}
p^{(K)}(x\mid c)
\;\propto\;
p^{(0)}(x\mid c)\;
\exp\!\left(\frac{K}{\beta}\, r(c,x)\right).
\end{equation}
Thus, under exact per-epoch optimization and a fixed reward $r(c,x)$, online training compounds the reward tilt:
the effective inverse temperature scales linearly with the number of epochs $K$.

\subsection{Concentration and Reward Hacking}
\cref{eq:app_online_tilting_unrolled} implies that as $K\to\infty$, the distribution $p^{(K)}(\cdot\mid c)$
concentrates its mass on the set of maximizers of the reward for that prompt.
Let $r_{\max}(c)=\sup_x r(c,x)$ and define the maximizer set
$S(c)=\{x:\; r(c,x)=r_{\max}(c)\}$.
Under mild regularity assumptions (e.g., $p^{(0)}(\cdot\mid c)$ assigns nonzero mass near $S(c)$),
the sequence $\{p^{(K)}(\cdot\mid c)\}$ becomes increasingly peaked on $S(c)$.
In practice, when $r(c,x)$ is produced by an imperfect reward model, this progressive sharpening can amplify reward
exploitation, i.e., \emph{reward hacking}.

\subsection{Remarks on EMA References}
If the reference model is updated via EMA in parameter space, the induced recursion over distributions is not exactly the
multiplicative update in \cref{eq:app_online_tilting_recursion}.
Nevertheless, EMA typically interpolates between keeping the reference fixed and fully replacing it with the current student,
effectively reducing the rate at which the tilt coefficient grows with epoch.
The qualitative conclusion remains: without an explicit anchoring term to the initial reference, repeated online updates tend
to accumulate reward tilt and can lead to overly concentrated policies that are more prone to reward hacking.

\section{GVPO and Reward Distill as Special Cases}
\label{supp:special_case}

\paragraph{Extreme case: \(K=2\) and \(\tau\to 0\) recovers reward distillation.}
Consider two samples \(\{x^1,x^2\}\) with rewards \(r_1=r(c,x^1)\) and \(r_2=r(c,x^2)\),
and assume w.l.o.g. \(r_1\ge r_2\). As \(\tau\to 0\), the softmax becomes one-hot:
\[
w_1\to 1,\qquad w_2\to 0.
\]
Hence \(\overline{r}_w\to r_1\) and \(\overline{R_\theta}_w\to R_1\) where
\(R_i=R_\theta(c,x^i)\). The centered terms satisfy
\[
\Delta_{r,w}^1\to 0,\quad \Delta_{r,w}^2\to r_2-r_1,
\qquad
\Delta_{R,w}^1\to 0,\quad \Delta_{R,w}^2\to R_2-R_1.
\]
Plugging into ~\cref{eq:rwrd_loss} (uniformly averaging over
\(i\in\{1,2\}\)) yields
\begin{align*}
\mathcal{L}_{\mathrm{distill\_GVPO}}^{(\tau\to 0)}
&=
\mathbb{E}_{\rho(c,x^1,x^2)}
\left[
\frac{1}{2}\Big( (0-0)^2 + \big((r_2-r_1)-(R_2-R_1)\big)^2 \Big)
\right] \\
&=
\mathbb{E}_{\rho(c,x^1,x^2)}
\left[
\frac{1}{2}\Big( (r_1-r_2)-(R_1-R_2)\Big)^2
\right].
\end{align*}
This is exactly the reward distillation loss proposed in \cite{mao2024don,gao2024rebel,fisch2024robust}.

\paragraph{Extreme case: \(\tau\to \infty\) recovers GVPO~\cite{zhang2025gvpo}.}
When \(\tau\to \infty\), the softmax weights become uniform:
\begin{equation}
w_i(c;\tau) \;=\; \frac{\exp\!\left(r(c,x_i)/\tau\right)}{\sum_{j=1}^K \exp\!\left(r(c,x_j)/\tau\right)}
\;\xrightarrow[\tau\to\infty]{}\;
\frac{1}{K}.
\end{equation}
Therefore the weighted means reduce to simple averages,
\begin{equation}
\overline{r}_w(c,\{x_i\}) \xrightarrow[\tau\to\infty]{} \frac{1}{K}\sum_{j=1}^K r(c,x_j),
\qquad
\overline{R_\theta}_w(c,\{x_i\}) \xrightarrow[\tau\to\infty]{} \frac{1}{K}\sum_{j=1}^K R_\theta(c,x_j),
\end{equation}
and consequently we show that GVPO is recovered as a special case of the proposed framework.

\section{Ratio-based Reward Distillation and Connection to InfoNCA}
\label{supp:ratio_based}

Unlike in the method section in the main text where we use the difference-based relationship to construct the loss, we can also construct the loss using ratio-based relationship.

~\cref{eq:rwrd_boltzmann} implies that, up to the (unknown) partition function \(Z(c)\), the exponentiated reward induces an energy-based density over responses proportional to a power of the optimal density ratio. Concretely, rewriting in terms of a probability ratio yields
\begin{equation}\label{eq:reward_and_density_ratio2}
\frac{\exp(r(c,x))}{Z(c)}
\;=\;
\left(\frac{p_{\theta^*}(x\mid c)}{p_{\mathrm{ref}}(x\mid c)}\right)^{\beta}.
\end{equation}
This suggests viewing both the true reward \(r\) and the model-induced reward \(r_\theta\) as defining \emph{energies} over a finite candidate set. Given a context \(c\) and a set of candidates \(\{x_i\}_{i=1}^K\), introduce a generic nonnegative weighting scheme \(\{w_i\}_{i=1}^K\) (e.g., importance weights, sampling correction, or uniform weights) and define the corresponding normalized distributions
\begin{equation}\label{eq:weighted_softmax_defs}
\begin{aligned}
q^*(i\mid c,\{x\})
&\;\triangleq\;
\frac{w_i\,\exp\!\big(r(c,x_i)/\beta\big)}{\sum_{j=1}^K w_j\,\exp\!\big(r(c,x_j)/\beta\big)},\\
q_\theta(i\mid c,\{x\})
&\;\triangleq\;
\frac{w_i\,\exp\!\big(r_\theta(c,x_i)/\beta\big)}{\sum_{j=1}^K w_j\,\exp\!\big(r_\theta(c,x_j)/\beta\big)}.
\end{aligned}
\end{equation}
When \(r_\theta=r\) (up to an additive constant independent of \(x\)), these two distributions coincide:
\begin{equation}\label{eq:reward_and_density_ratio3}
q^*(i\mid c,\{x\})
=
q_\theta(i\mid c,\{x\}).
\end{equation}
Crucially, the normalization in ~\cref{eq:weighted_softmax_defs} removes dependence on the unknown \(Z(c)\), converting reward matching into distribution matching over a finite set.

\paragraph{Divergence minimization view.}
A natural objective is therefore to fit \(q_\theta\) to \(q^*\) by minimizing a divergence over sets:
\begin{equation}\label{eq:dist_match_general}
\mathcal{L}_{\mathrm{dist}}(\theta)
\;=\;
\mathbb{E}_{\rho(c,\{x_i\})}\Big[ D\!\big(q^*(\cdot\mid c,\{x\})\;\|\; q_\theta(\cdot\mid c,\{x\})\big)\Big],
\end{equation}
where \(D(\cdot\|\cdot)\) can be chosen as forward KL, reverse KL, Jensen--Shannon, or other proper scoring rules.

\paragraph{Special case: uniform weights and forward KL recover InfoNCA.}
For the common choice \(w_i\equiv 1\), both \(q^*\) and \(q_\theta\) are categorical distributions on \(\{1,\dots,K\}\) \cite{oord2018representation}. Taking \(D\) to be the forward KL gives
\begin{align}
\mathcal{L}_{\mathrm{InfoNCA}}(\theta)
&\triangleq
\mathbb{E}_{\rho(c,\{x_i\})}\Big[
\mathrm{KL}\!\big(q^*(\cdot\mid c,\{x\})\;\|\; q_\theta(\cdot\mid c,\{x\})\big)
\Big] \nonumber\\
&=
\mathbb{E}_{\rho(c,\{x_i\})}\Big[
-\sum_{i=1}^K q^*(i\mid c,\{x\}) \log q_\theta(i\mid c,\{x\})
\Big] \;+\; \text{const.} \label{eq:infonca_ce}
\end{align}
Thus, minimizing forward KL is equivalent to minimizing cross-entropy from the ``teacher'' distribution induced by \(r\) to the ``student'' distribution induced by \(r_\theta\).

\section{Additional Results} 
\label{sec:app_additional_result}

\begin{figure}[t!]
    \centering
    \begin{minipage}[t]{0.9\textwidth}
    \centering
    \begin{subfigure}[t!]{0.48\textwidth}
        \centering
        \includegraphics[width=\textwidth]{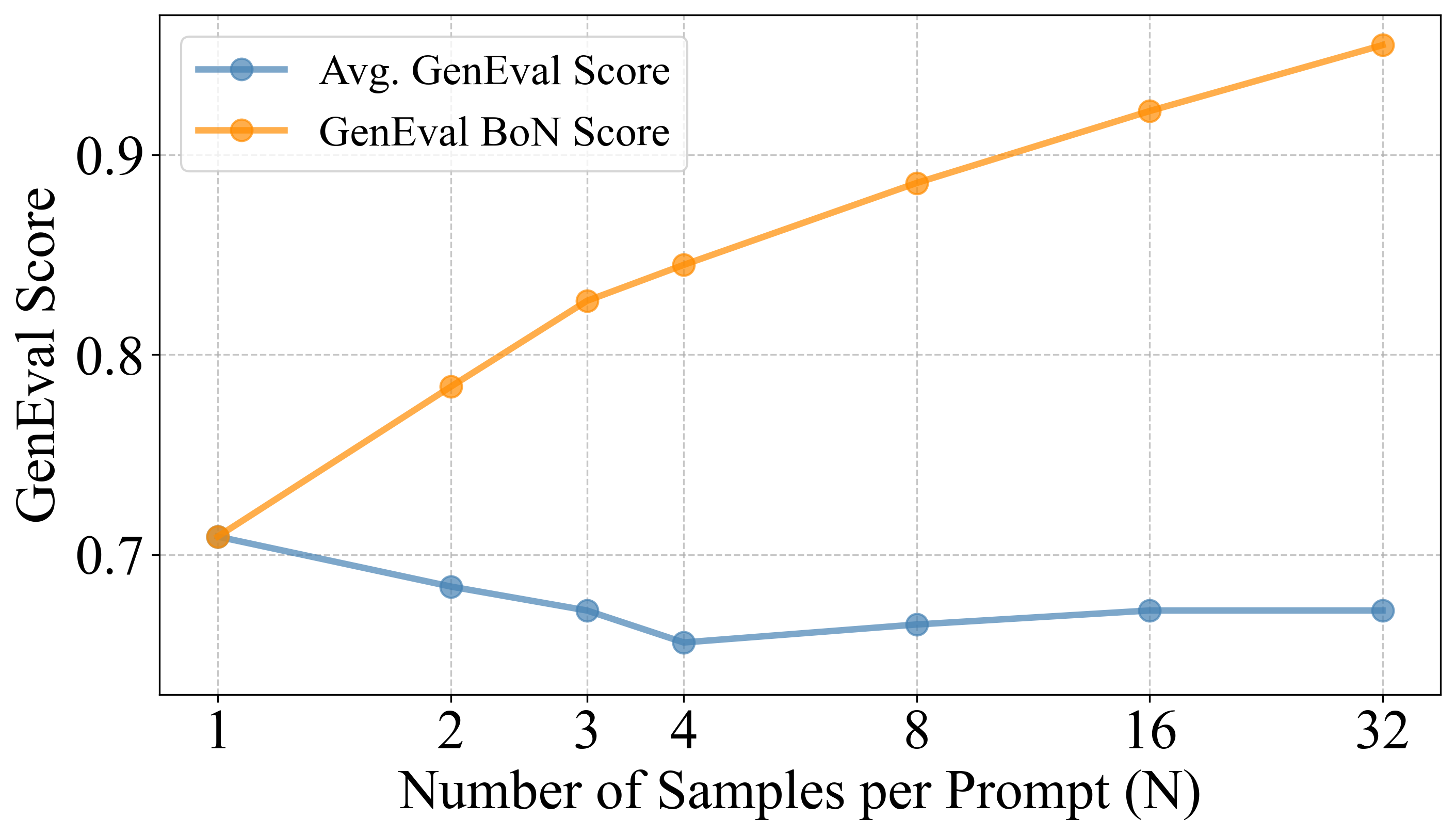}
        \caption{BoN results with SD3.5-M}
    \end{subfigure}
    \hfill
    \begin{subfigure}[t!]{0.48\textwidth}
        \centering
        \includegraphics[width=\textwidth]{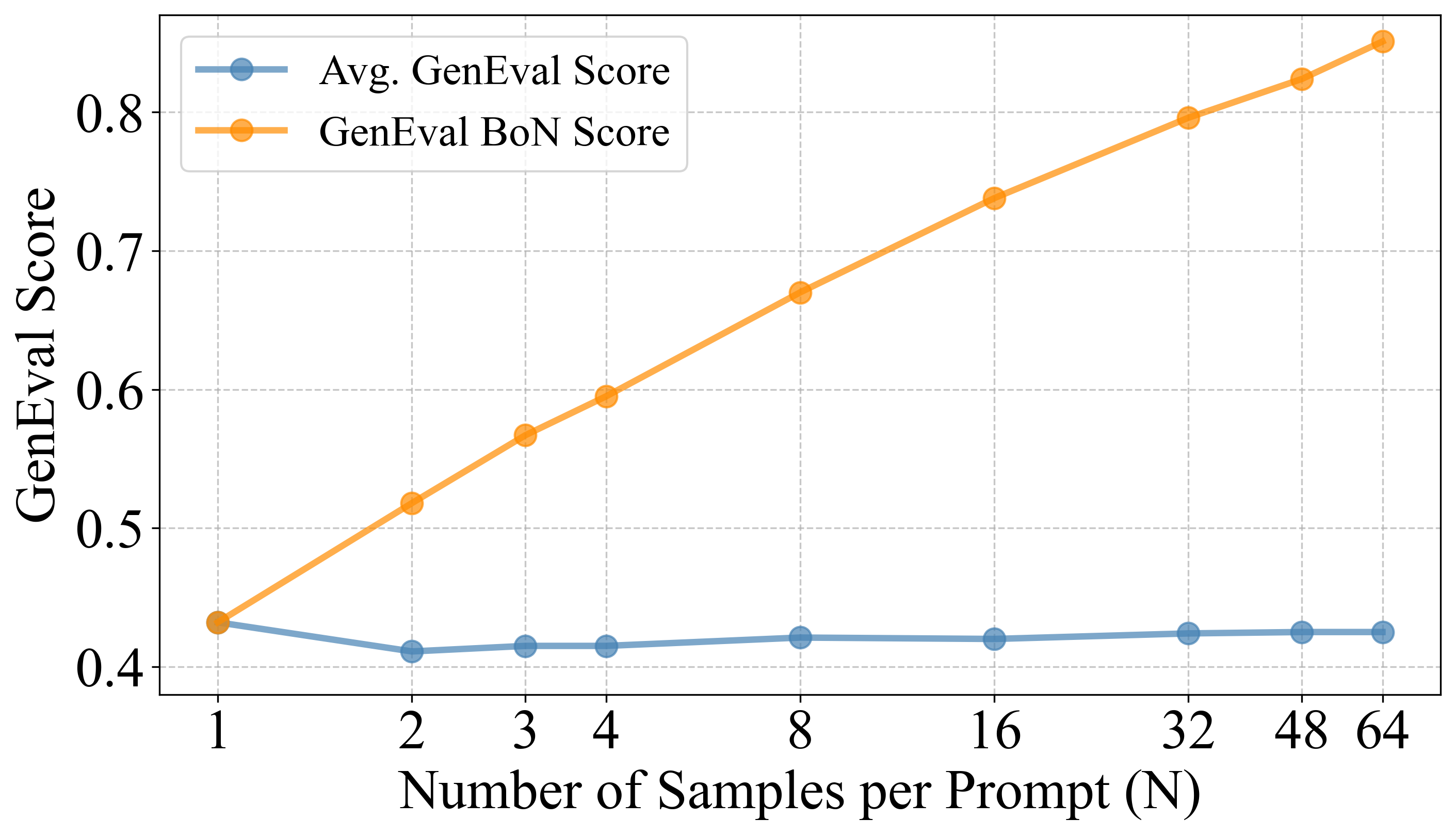}
        \caption{BoN results with SD1.5}
    \end{subfigure}
    \caption{BoN results. The orange curve is the BoN performance, and the blue curve is the average performance of the current $N$ samples.}
    \label{fig:bon_reference}
    \end{minipage}
\end{figure}

\subsection{Best-of-N (BoN) Performance of SD3.5-M and SD1.5}
\label{sec:BoN}
{Best-of-N (BoN) is an inference-time scaling strategy that generates $N$ independent samples and selects the one with the highest reward score \cite{gao2023scaling,beirami2024theoretical,eisenstein2023helping,gui2024bonbon,ma2025inference}. }
We report the BoN performance of SD3.5-M and SD1.5 on \texttt{GenEval} in \cref{fig:bon_reference} as a reference.
{The orange curve reports BoN performance as $N$ increases (log-linear for both SD3.5-M and SD1.5), while the blue curve shows the average reward score on the current $N$ samples.}
The results suggest that RL fine-tuning can effectively amortize this inference-time cost, matching or exceeding BoN performance without requiring multiple samples at test-time.

\subsection{Multi-reward Training}
\label{appd:multi_reward}

\begin{figure}[t!]
    \centering
    \includegraphics[width=0.95\linewidth]{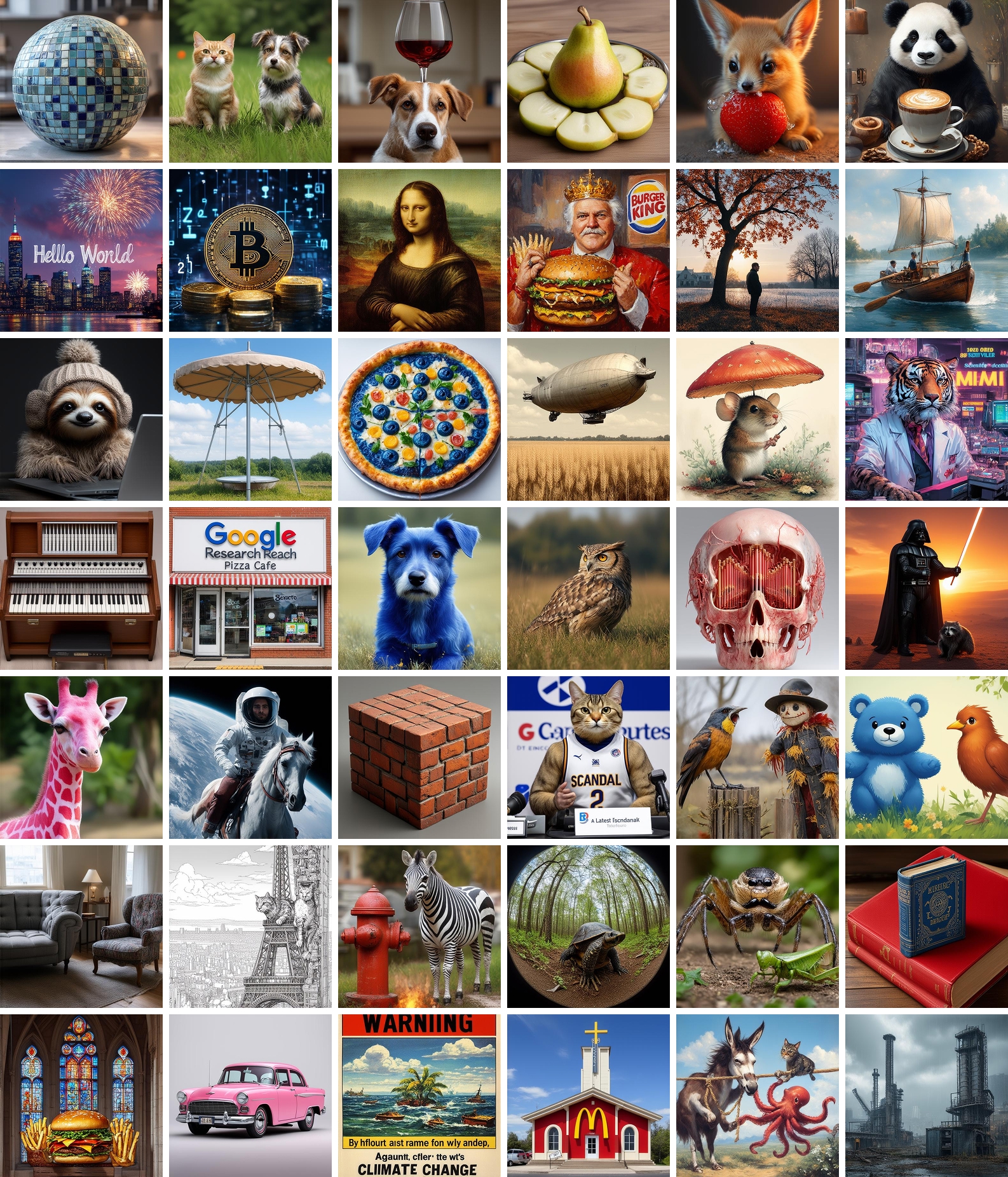}
    \caption{{Qualitative visual generations} from our model trained with multiple rewards. The text prompts are selected from the \texttt{DrawBench}.}
    \label{fig:visual_multi_reward}
\end{figure}

We also evaluate our method under a multi-reward training setting, where the model is jointly optimized
using multiple rewards. Following prior work~\cite{zheng2025diffusionnft, choi2026rethinking},
we report \texttt{PickScore}, \texttt{CLIPScore}, and \texttt{HPSv2.1} as task metrics and trained on the \texttt{PickScore} prompt dataset, alongside \texttt{Aesthetics} and \texttt{ImageReward}
(\texttt{ImgRwd}) as preference scores, all evaluated on \texttt{DrawBench}~\cite{saharia2022photorealistic} prompts.
As shown in \cref{tab:multi_reward}, CRD achieves competitive performance across all metrics
compared to baselines such as Flow-GRPO~\cite{liu2025flow} and Choi~\textit{et al.}~\cite{choi2026rethinking}.
Notably, applying CFG sampling further improves \texttt{CLIPScore}, suggesting better text alignment,
although at a slight cost to aesthetic quality at higher CFG values.
Corresponding visualizations can be found in \cref{fig:visual_multi_reward}.

\begin{table*}[t]
    \centering
    \caption{\textbf{Performance on Human Preference benchmarks with Multi-reward Training.} Task metrics, image quality and preference scores are all evaluated on \texttt{DrawBench} \cite{saharia2022photorealistic} prompts. Baseline results are taken from Flow-GRPO~\cite{liu2025flow} and \cite{choi2026rethinking}, or reimplemented when unavailable in the original papers. Flow-GRPo is trained on \texttt{PickScore} only. \texttt{ImgRwd}: \texttt{ImageReward}.}
    \vspace{-10pt}
    \resizebox{0.9\linewidth}{!}{
        \begin{tabular}{lccccccc}
            \toprule
            \multirow{2}{*}{\textbf{Model}} & \multicolumn{3}{c}{\textbf{Task Metric}} & \multicolumn{2}{c}{\textbf{Preference Score}}    \\ \cmidrule(lr){2-4} \cmidrule(lr){5-6} 
                                   & \textbf{\texttt{PickScore}} $\uparrow$ & \textbf{\texttt{CLIPScore}} $\uparrow$ & \textbf{\texttt{HPSv2.1}} $\uparrow$ & \textbf{\texttt{Aesthetics}} $\uparrow$& \textbf{\texttt{ImgRwd}} $\uparrow$ \\ \midrule
            SD3.5-M & 22.34 & 27.99 & 0.279 & 5.39 & 0.87  \\
            \midrule
            \multicolumn{6}{c}{\textit{Visual Text Rendering}} \\
            \midrule
            Flow-GRPO~\cite{liu2025flow} (w/o KL)  & 23.41 & — & — & 6.15 & 1.24 \\
            Flow-GRPO~\cite{liu2025flow} (w/ KL)  & 23.31 & 27.81 & 0.315 & 5.92 & 1.28 \\
            DiffusionNFT~\cite{zheng2025diffusionnft} & 23.61 & 28.80 & 0.344 & 6.04 & 1.46 \\
            
            Choi \etal~\cite{choi2026rethinking} & 23.68 & 29.60 & 0.325 & 6.06 & 1.45 \\
            \rowcolor{gray!20} \method{} & 23.27 & 29.63 & 0.321 & 5.70 & 1.35 \\
            \rowcolor{gray!20} \qquad + CFG sampling=1.5 & 23.22 & 29.91 & 0.323 & 5.60 & 1.39 \\
            \rowcolor{gray!20} \qquad + CFG sampling=3.0 & 22.89 & 30.01 & 0.315 & 5.47 & 1.36 \\
            \bottomrule
            \end{tabular}
}
\label{tab:multi_reward}
\end{table*}

\subsection{Adaptive KL Regularization on DiffusionNFT}
\label{appd:kl_nft}

\begin{figure}[t!]
    \centering
    \begin{minipage}[t]{0.95\textwidth}
        \centering
        \begin{subfigure}[t]{0.44\textwidth}
            \centering
            \includegraphics[width=\textwidth]{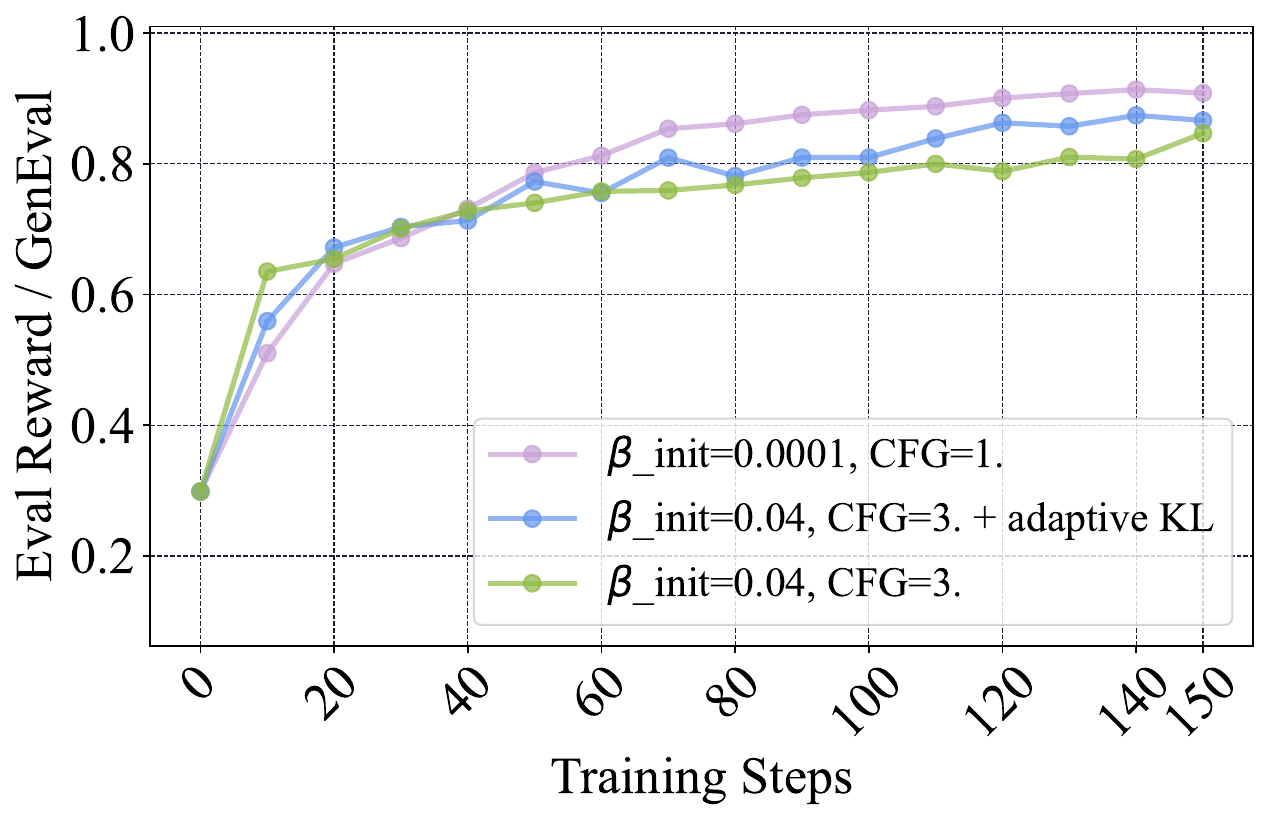}
            \caption{Evaluation \texttt{GenEval} score during training, under different $\beta_{\mathrm{init}}$ and CFG values.}
            \label{fig:nft_kl_curve}
        \end{subfigure}
        \hfill
        \begin{subfigure}[t]{0.55\textwidth}
            \centering
            \includegraphics[width=\textwidth]{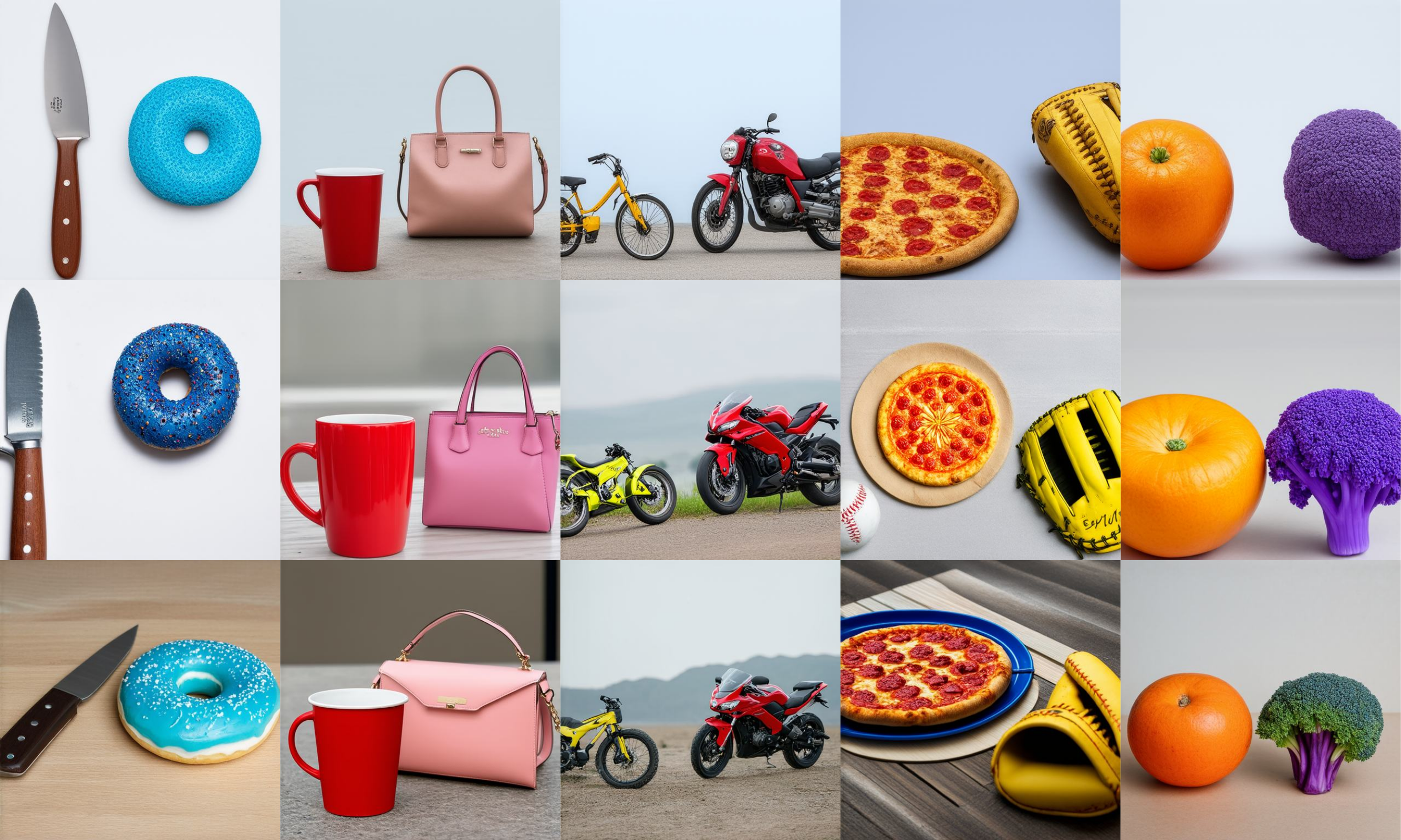}
            \caption{Visual comparison, from top to bottom corresponding to the \textcolor[HTML]{C8A0D8}{pink}, \textcolor[HTML]{6495ED}{blue} and \textcolor[HTML]{8EBA42}{green} curve in \cref{fig:nft_kl_curve}.}
        \end{subfigure}
        \caption{Effectiveness of adaptive initial KL on DiffusionNFT.}
        \label{fig:nft_kl}
    \end{minipage}
\end{figure}

In \cref{fig:nft_kl}, we demonstrate that the proposed Adaptive KL Regularization also effectively mitigates reward hacking when applied to DiffusionNFT, preserving visual quality (fewer white backgrounds) while maintaining high reward scores.

\subsection{Additional Ablations}

\begin{figure}[t!]
    \centering
    \begin{minipage}[t]{0.64\textwidth}
        \centering
        \begin{subfigure}[t]{0.48\textwidth}
            \centering
            \includegraphics[width=\textwidth]{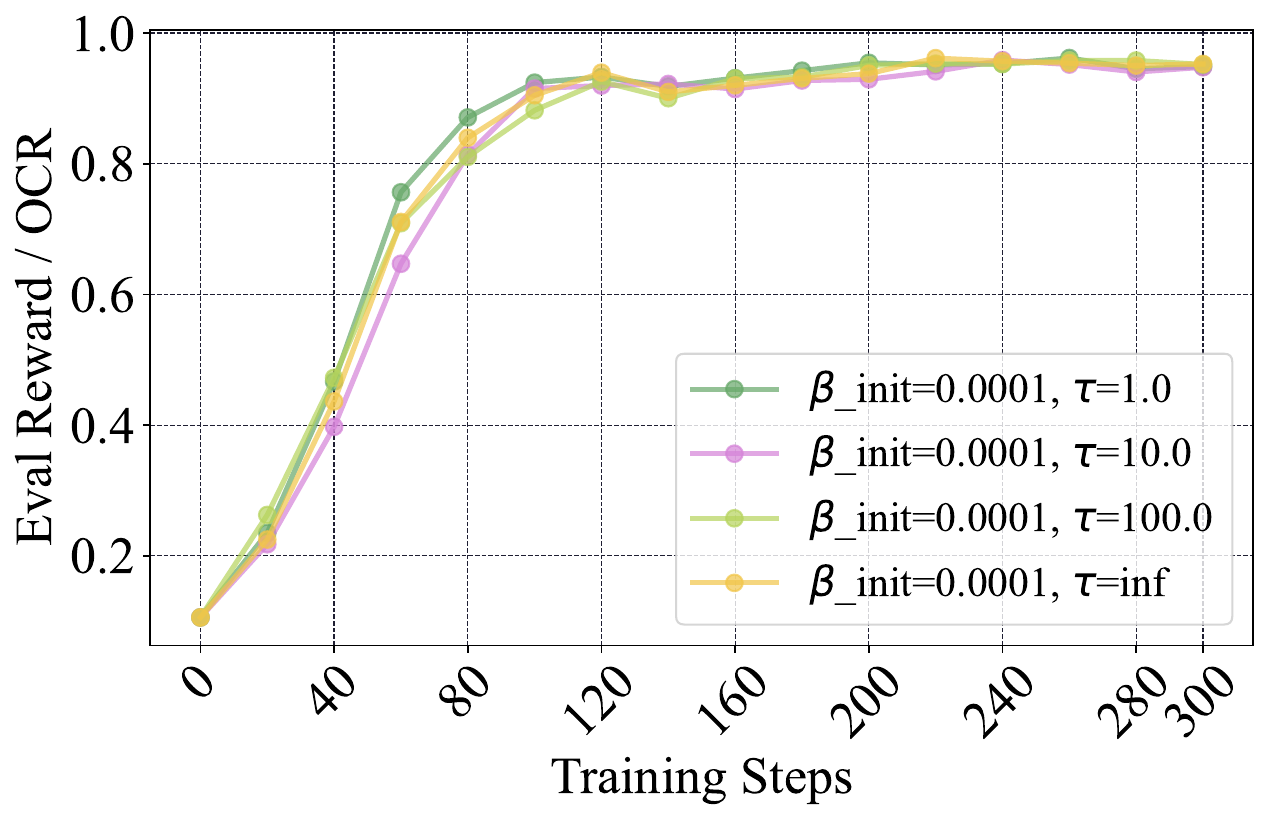}
            \caption{$\beta_{\mathrm{init}}$=0.0001; CFG=1.0}
        \end{subfigure}
        \hfill
        \begin{subfigure}[t]{0.48\textwidth}
            \centering
            \includegraphics[width=\textwidth]{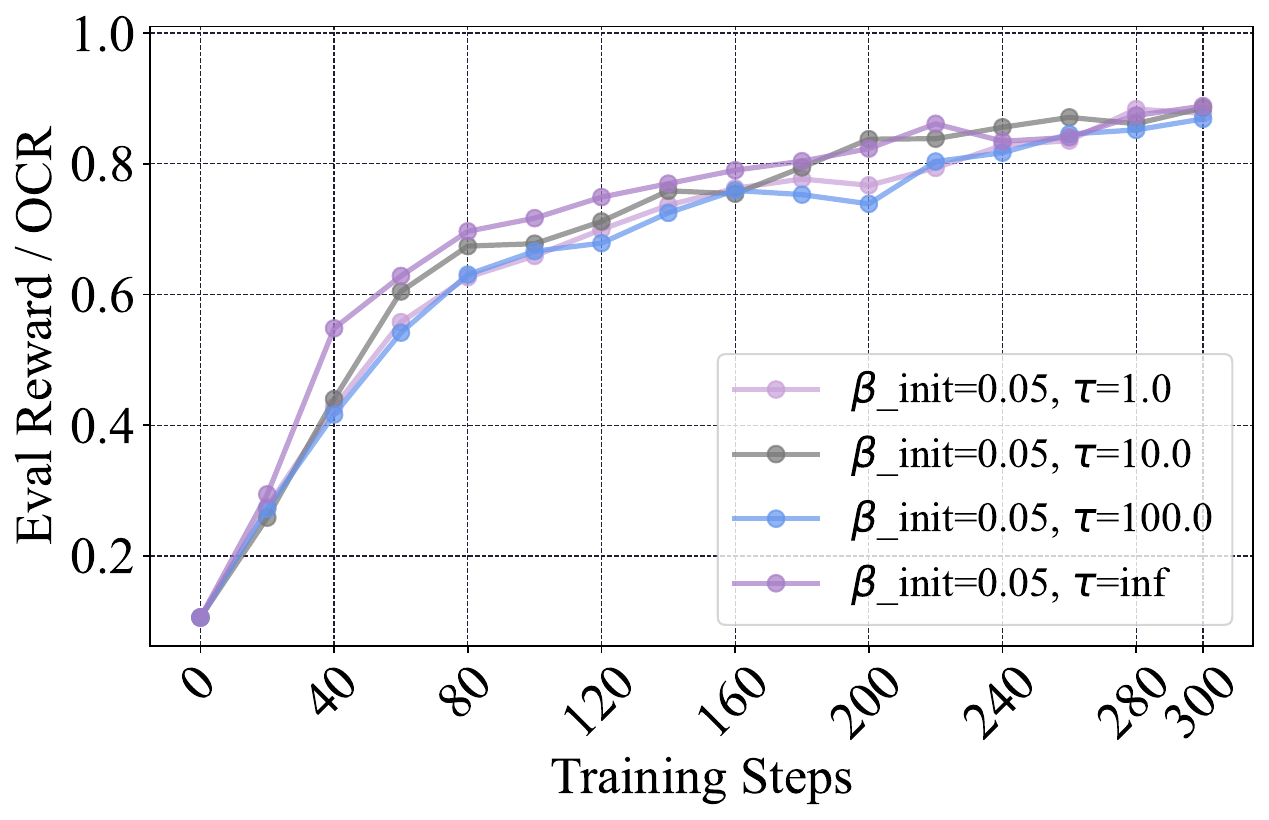}
            \caption{$\beta_{\mathrm{init}}$=0.05; CFG=3.0}
        \end{subfigure}
        \caption{Ablations on different temperatures $\tau$}
        \label{fig:temp_ablate}
    \end{minipage}
    \hfill
    \begin{minipage}[t]{0.32\textwidth}
        \centering
        \includegraphics[width=\textwidth]{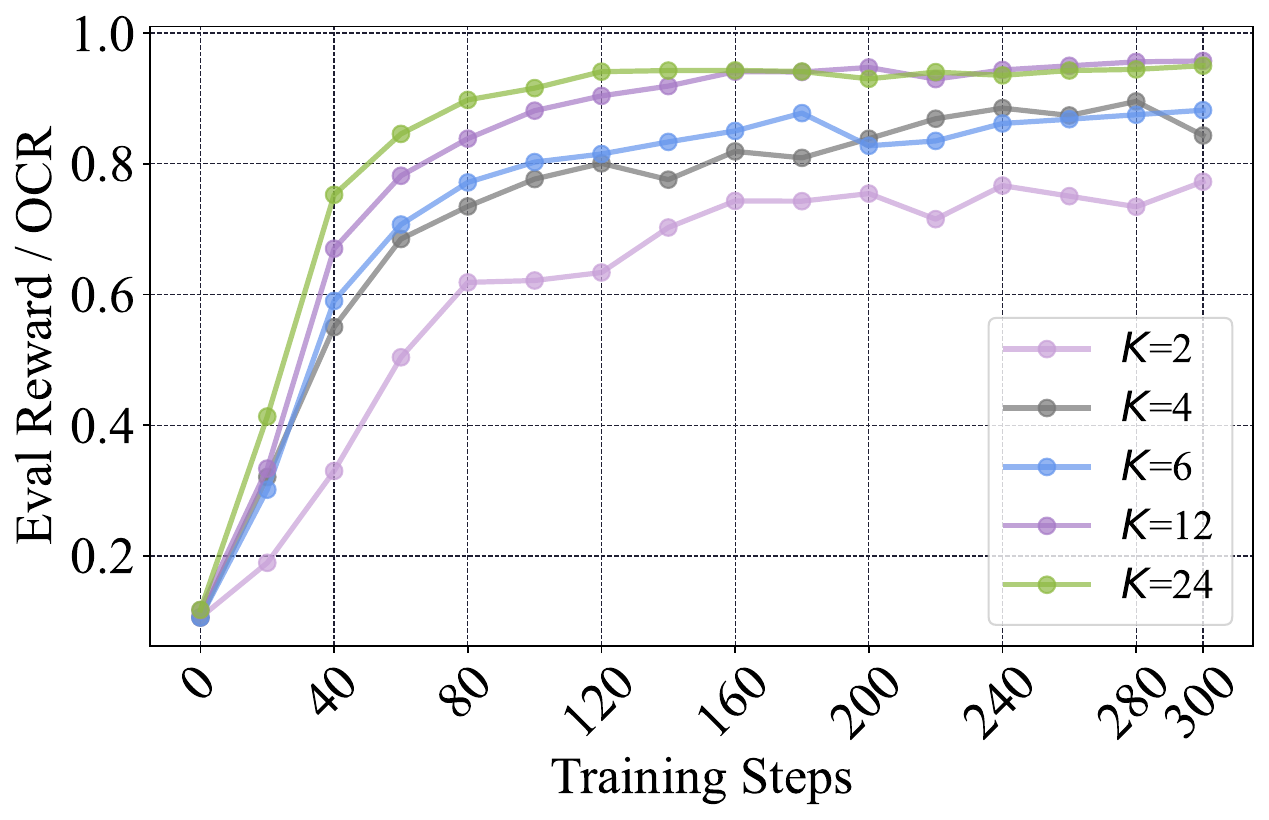}
        \caption{Ablation on group size $K$ with OCR reward}
        \label{fig:K_ablate}
    \end{minipage}
\end{figure}

\paragraph{Temperature in \method{} objective.}
As shown in \cref{fig:temp_ablate}, varying $\tau$ over a wide range (1.0, 10.0, 100.0, $\infty$) yields remarkably similar reward curves across both small and large $\beta_{\mathrm{init}}$, suggesting that \method{} is largely insensitive to the choice of $\tau$. For simplicity, we use uniform weighting (\ie $\tau{=}\infty$) in all other experiments.

\paragraph{Group size $K$.}
\cref{fig:K_ablate} shows a clear trend: larger group sizes lead to faster convergence and higher final rewards, with $K{=}24$ achieving near-perfect OCR scores by the end of training, as larger groups provide richer gradient signal per update. Notably, even $K{=}2$ yields meaningful learning, demonstrating the robustness of \method{} under minimal sampling budgets.

\begin{figure}[t!]
    \centering
    \begin{subfigure}[t]{0.32\textwidth}
        \centering
        \includegraphics[width=\textwidth]{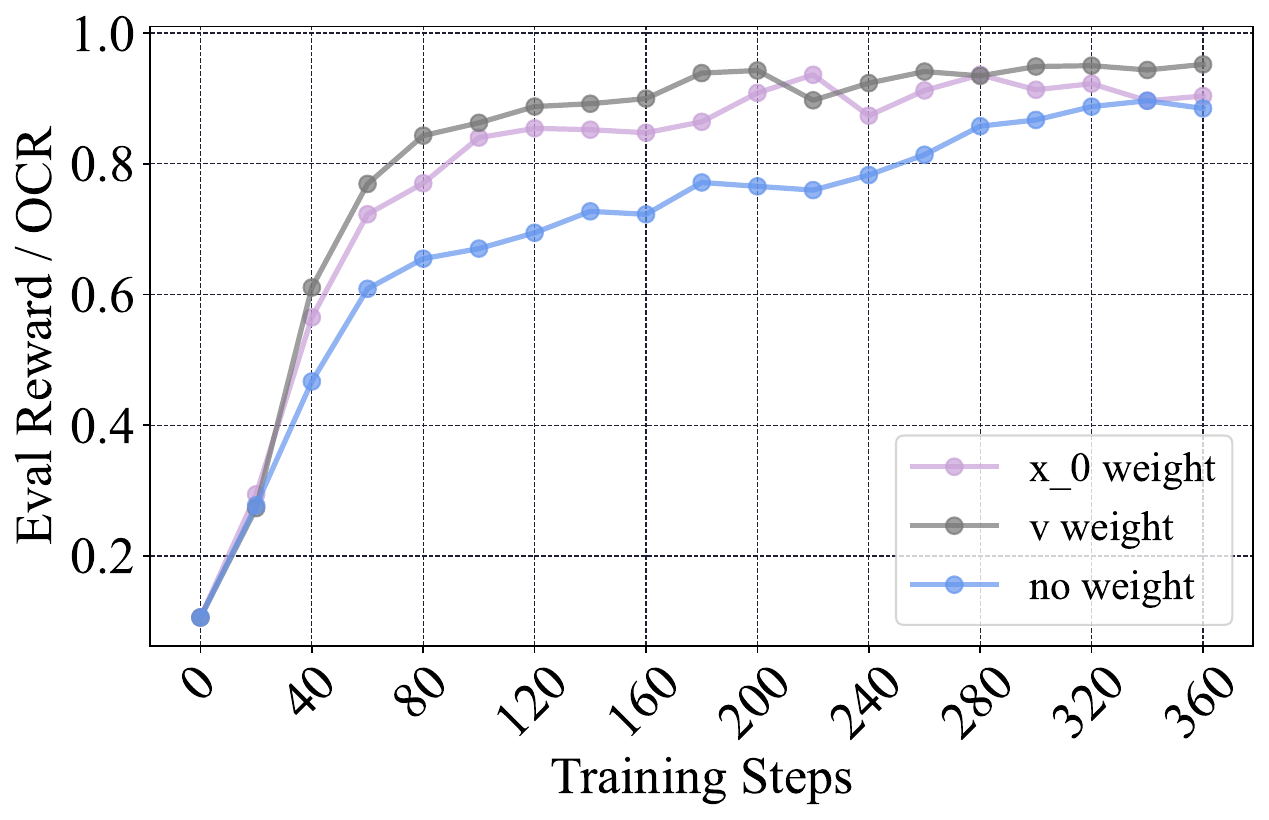}
        \caption{Evaluation reward of the current model $\theta$}
    \end{subfigure}
    \hfill
    \begin{subfigure}[t]{0.32\textwidth}
        \centering
        \includegraphics[width=\textwidth]{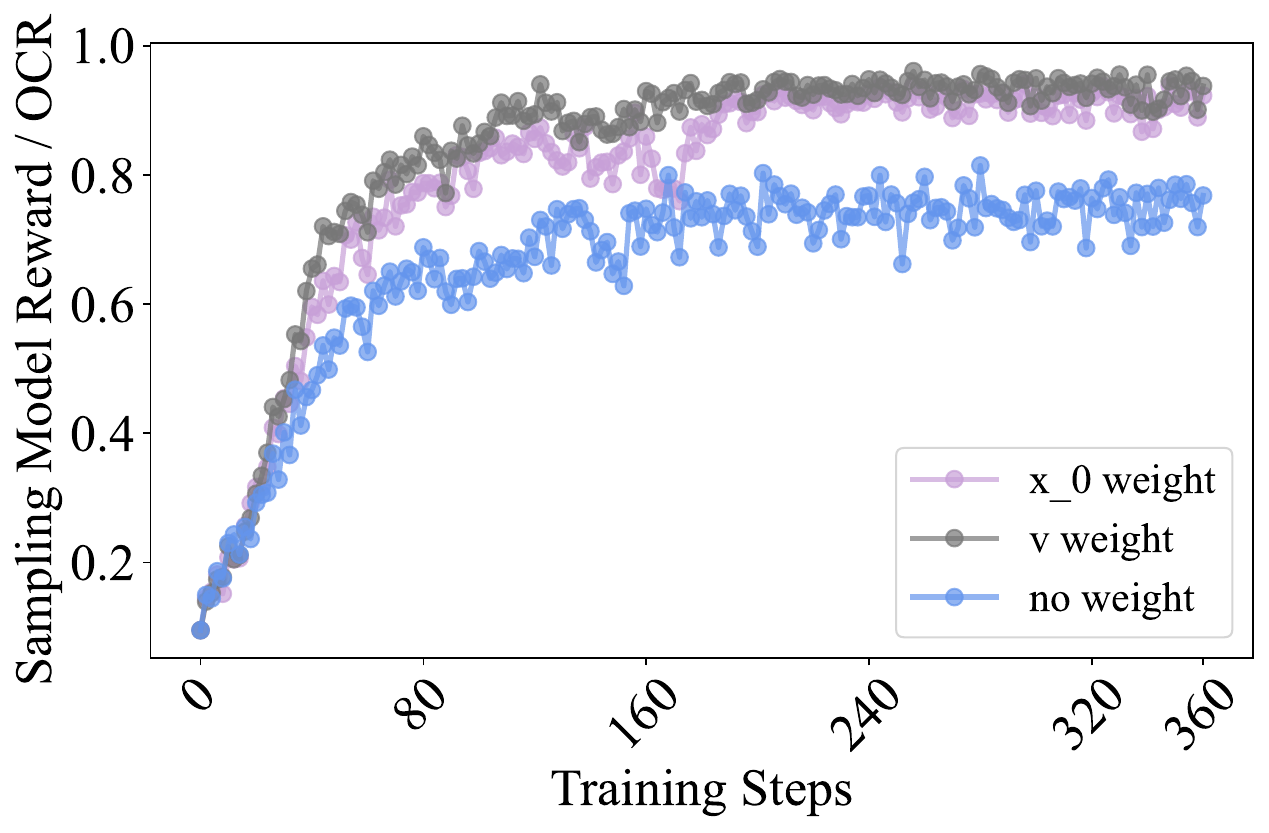}
        \caption{Reward value during training of the sampling model $p_{\mathrm{samp}}$}
    \end{subfigure}
    \hfill
    \begin{subfigure}[t]{0.32\textwidth}
        \centering
        \includegraphics[width=\textwidth]{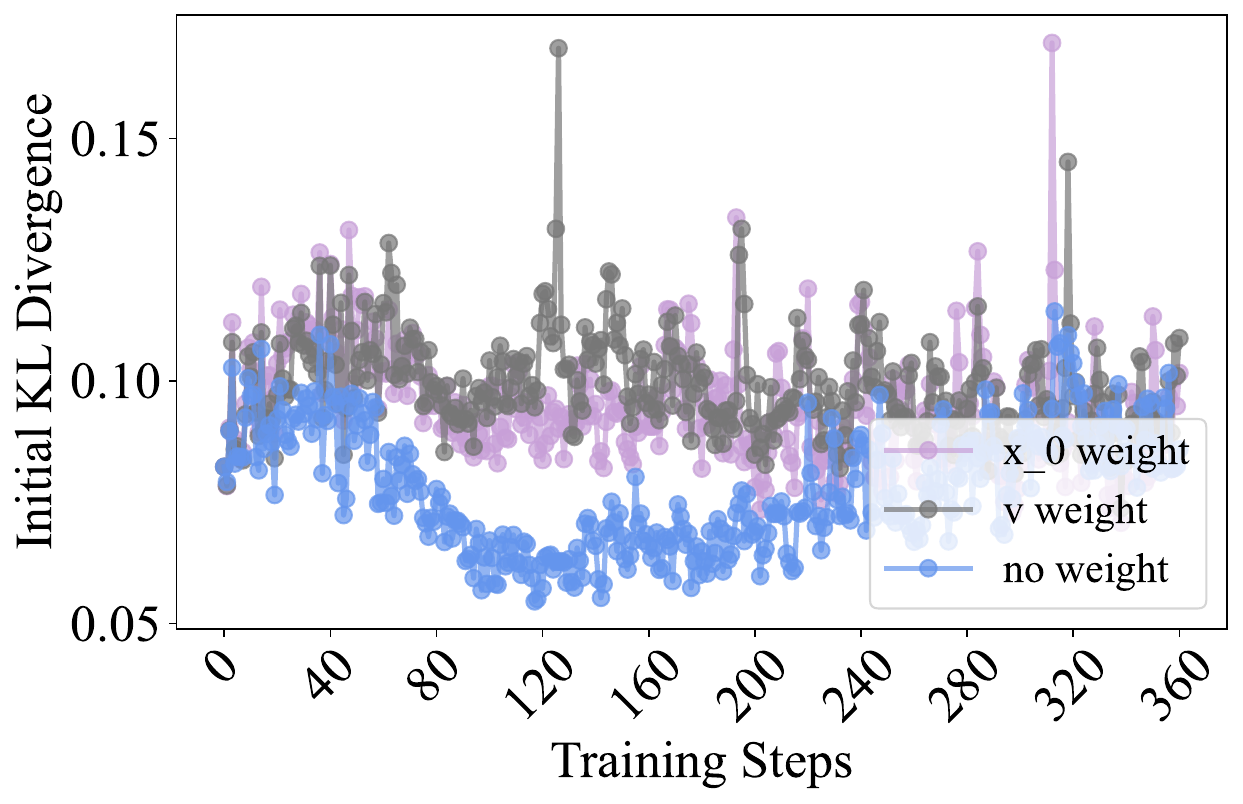}
        \caption{Initial KL value during training}
    \end{subfigure}
    \caption{Ablations on adaptive weighting in ELBO estimator
    }
    \label{fig:adap_w_ablate}
\end{figure}

\paragraph{Adaptive weighting in ELBO estimator.}
The original adaptive weighting~\cite{yin2024one,zheng2025diffusionnft} is built on the $x_0$ representation, which introduces an extra coefficient $t$ compared to the estimator in \cref{eq:rwrd_policy_reward_elbo_adaptive}:
\begin{equation}
\label{eq:rwrd_policy_reward_elbo_adaptive_x0}
\widehat{R}_\theta^{\mathrm{adap}\_{x_0}}(c,x)
\;\triangleq\;
-\beta\,
\mathbb{E}_{t,\epsilon}\!\left[
\frac{td\left\|v_\theta(x_t,t\mid c)-v_\mathrm{target}\right\|_2^2}{\mathrm{sg}(\left\|v_\theta(x_t,t\mid c)-v_\mathrm{target}\right\|_1)}
-
\frac{td\left\|v_{\mathrm{ref}}(x_t,t\mid c)-v_\mathrm{target}\right\|_2^2}{\mathrm{sg}(\left\|v_{\mathrm{ref}}(x_t,t\mid c)-v_\mathrm{target}\right\|_1)}
\right].
\end{equation}
As shown in \cref{fig:adap_w_ablate}, our $v$-prediction adaptive weighting (\cref{eq:rwrd_policy_reward_elbo_adaptive}) achieves the highest evaluation reward and sampling model reward, while maintaining stable KL divergence throughout training. 
We therefore adopt our adaptive weighting scheme in all experiments.

\subsection{Additional Qualitative Results}

\label{sec:app_quali}

\paragraph{Visualization during training process}
We visualize generated samples at successive checkpoints throughout training, starting from step 0 and sampled at every 200-step interval. As training progresses, the model learns to render text more accurately within the generated images without collapsing, reflecting steady improvements in text fidelity over time.

\begin{figure}[t!]
    \centering
    \includegraphics[width=0.92\linewidth]{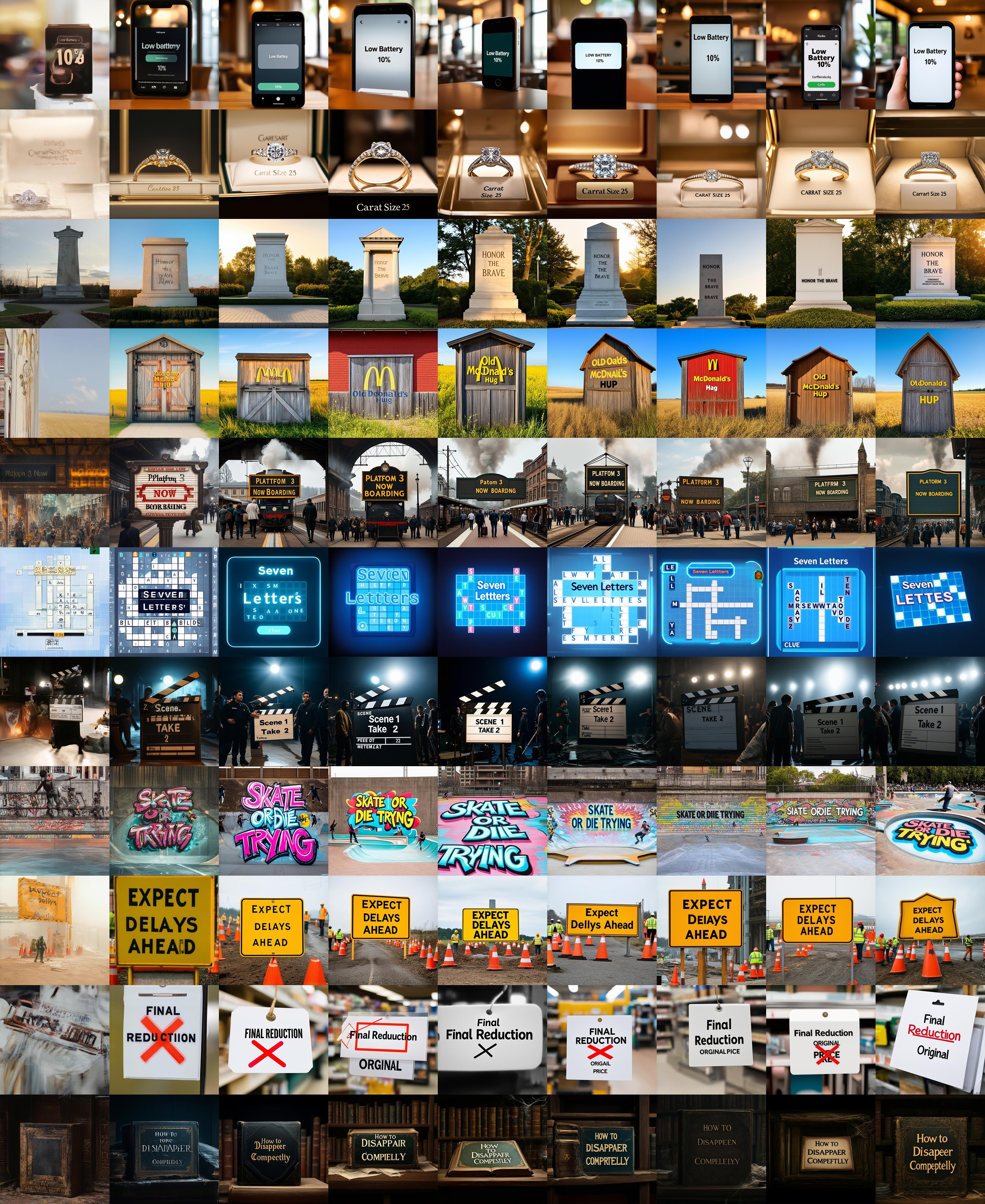}
    \noindent\makebox[\linewidth]{%
        \raisebox{0pt}[0pt][0pt]{$\xrightarrow{\hspace{0.85\linewidth}}$}%
    }\\[-4pt]
    \noindent\makebox[\linewidth]{\small Training Progress}
    \caption{We visualize samples generated {with our method} at successive checkpoints (from left to right) throughout training, starting from step 0 and sampled at every 200-step interval.}
    \vspace{-15pt}
    \label{fig:evolution}
\end{figure}

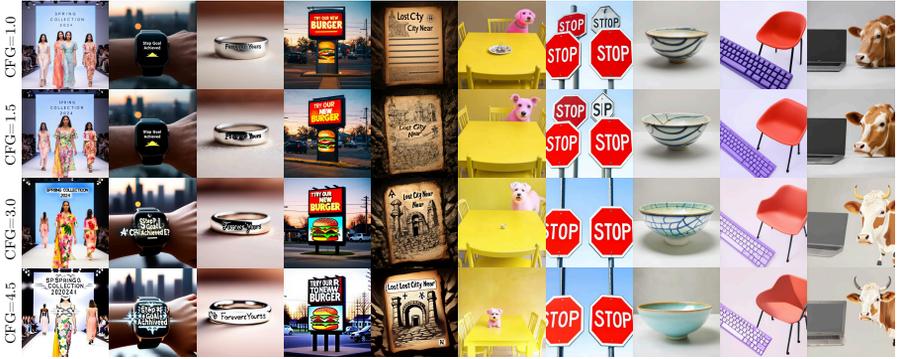
\begin{figure}[t!]
\centering
\begin{overpic}[width=0.98\linewidth]{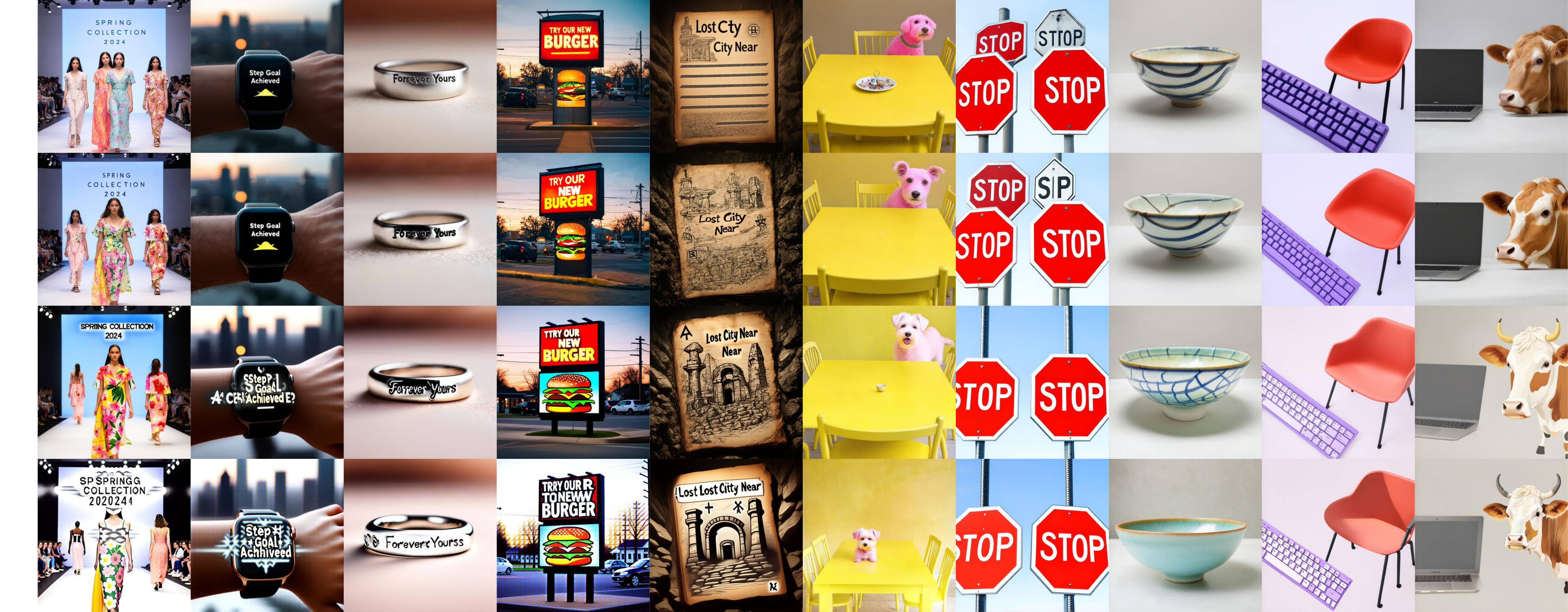}%
\put(0.5,30.0){\rotatebox{90}{\color{black}\scalebox{0.6}{ CFG=1.0}}}
\put(0.5,20.4){\rotatebox{90}{\color{black}\scalebox{0.6}{ CFG=1.5}}}
\put(0.5,10.){\rotatebox{90}{\color{black}\scalebox{0.6}{ CFG=3.0}}}
\put(0.5,0.8){\rotatebox{90}{\color{black}\scalebox{0.6}{ CFG=4.5}}}
\end{overpic}
\caption{
\small Visualization of generation with different inference CFG values. The left 5 columns are generated with prompts from \texttt{GenEval} and the right 5 columns are generated with prompts from \texttt{OCR} testsets, respectively.
}
\label{fig:visual_cfg}
\end{figure}

\paragraph{Inference with higher CFG}
In \cref{fig:visual_cfg}, we compare samples generated with increasing CFG scales. 
While moderate CFG typically improves prompt adherence, we observe a consistent degradation in visual fidelity as CFG becomes large. 
For text-centric prompts, higher CFG often introduces typographic artifacts such as broken glyphs, and unstable letter shapes, even when the overall image appears more strongly aligned with the prompt. 
For compositional prompts, increasing CFG does not reliably improve prompt following and can even reduce it. For example, the \emph{four} stop sign prompt frequently violates the counting constraint at the higher CFG. 
Finally, we observe a style shift at large CFG values, where the cow in the last column gradually transitions from a photorealistic appearance to a more stylized, illustration-like rendering with simplified textures and exaggerated features.

\begin{figure}[t!]
    \centering
    \includegraphics[width=0.95\linewidth]{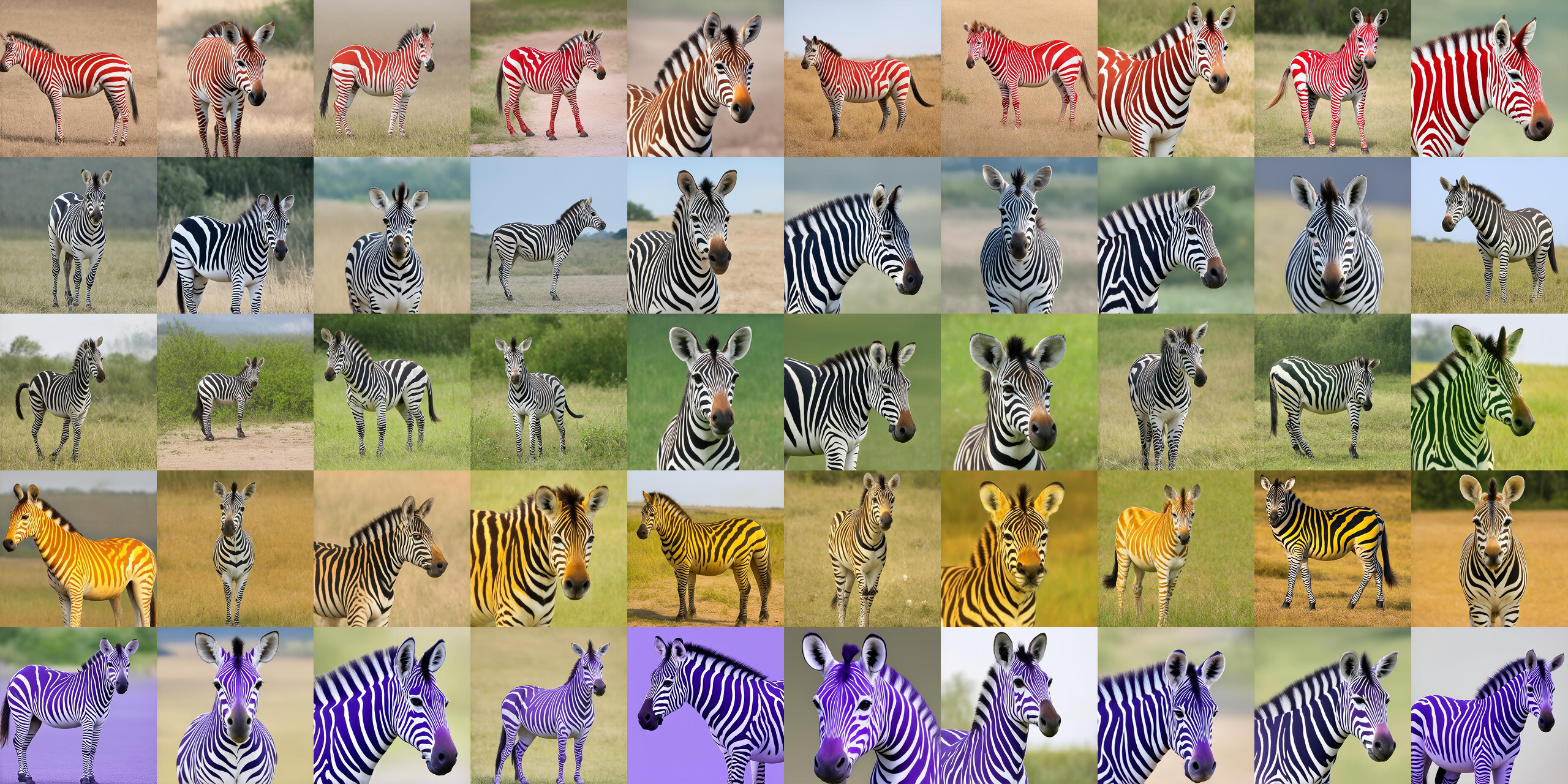}
    \caption{\textbf{Qualitative visual generations.} Each row shows generations from a different random seed for the same prompt. Prompts follow the template ``a photo of a \{\} zebra'', where the placeholder is replaced by the colors \textcolor{red}{red}, \textcolor{blue}{blue}, \textcolor{green}{green}, \textcolor{yellow}{yellow}, and \textcolor{purple}{purple} from top to bottom.}
    \label{fig:zebra}
\end{figure}

\paragraph{Colored zebra.}
We further demonstrate the model's ability to generate zebras in a variety of colors. The training dataset has a roughly uniform color distribution: 
\begin{center}
\small
\texttt{Counter(\{`white': 1857, `purple': 1817, `green': 1797, `brown': 1782, `black': 1771, `orange': 1749, `pink': 1742, `blue': 1722, `red': 1715, `yellow': 1696\}).}
\end{center}
Qualitative generations with simple prompts are shown in \cref{fig:zebra}. 
We highlight three observations: (\textit{1}) the model produces diverse and visually distinct outputs across colors; (\textit{2}) despite the balanced color distribution in training, the model struggles to generate convincing blue and green zebras (failure generation); and (\textit{3}) in most cases, the black stripes are replaced by the target color, with only rare cases where the zebra retains its black stripes alongside the assigned color.

\newpage

\end{document}